\documentclass[review]{elsarticle}

\usepackage{microtype}
\usepackage{graphicx}
\usepackage{subfigure}
\usepackage{booktabs}
\usepackage{amsmath}
\usepackage{amsfonts}
\usepackage[english]{babel}
\usepackage{amsthm}
\usepackage{nicefrac}
\usepackage{multicol}
\usepackage{multirow}
\usepackage[table]{xcolor}
\usepackage{adjustbox}
\usepackage{tablefootnote}
\usepackage{tikz}
\usepackage{makecell}
\usepackage{wrapfig}
\usepackage{bm}
\usepackage{natbib}
\usepackage{geometry}
\usepackage{hyperref}

\newtheorem{theorem}{Theorem}
\newtheorem{lemma}{Lemma}

\biboptions{numbers,sort&compress}

\journal{Neural Networks}

\begin{document}
\begin{frontmatter}

\title{Discovering Parametric Activation Functions}

\author[UT,Cognizant]{Garrett Bingham\corref{cor1}}
\ead{bingham@cs.utexas.edu}
\cortext[cor1]{Corresponding author}

\author[UT,Cognizant]{Risto Miikkulainen}
\ead{risto@cs.utexas.edu}

\affiliation[UT]{organization={The University of Texas at Austin},
            city={Austin},
            state={Texas},
            postcode={78712},
            country={USA}}

\affiliation[Cognizant]{organization={Cognizant AI Labs},
            addressline={649 Front St.},
            city={San Francisco},
            state={California},
            postcode={94111},
            country={USA}}

\begin{abstract}
Recent studies have shown that the choice of activation function can significantly affect the performance of deep learning networks. However, the benefits of novel activation functions have been inconsistent and task dependent, and therefore the rectified linear unit (ReLU) is still the most commonly used. This paper proposes a technique for customizing activation functions automatically, resulting in reliable improvements in performance. Evolutionary search is used to discover the general form of the function, and gradient descent to optimize its parameters for different parts of the network and over the learning process. Experiments with four different neural network architectures on the CIFAR-10 and CIFAR-100 image classification datasets show that this approach is effective. It discovers both general activation functions and specialized functions for different architectures, consistently improving accuracy over ReLU and other activation functions by significant margins.  The approach can therefore be used as an automated optimization step in applying deep learning to new tasks.
\end{abstract}


\end{frontmatter}

\usetikzlibrary{shapes.geometric, arrows, positioning, calc}
\tikzstyle{input} = [rectangle, rounded corners, text centered, draw=black, fill=red!30]
\tikzstyle{unary} = [rectangle, rounded corners, text centered, draw=black, fill=yellow!30]
\tikzstyle{binary} = [rectangle, rounded corners, text centered, draw=black, fill=blue!30]
\tikzstyle{arrow} = [thick, ->,>=stealth]
\tikzstyle{output} = [rectangle, rounded corners, text centered, draw=black, fill=green!30]
\tikzstyle{invisible} = [rectangle, text centered]
\tikzstyle{maybeparam} = [draw, circle, dashed, fill=cyan!30]

\section{Introduction}

The rectified linear unit ($\textrm{ReLU}(x) = \max\{x, 0\}$) is the most commonly used activation function in modern deep learning architectures \citep{nair2010rectified}.  When introduced, it offered substantial improvements over the previously popular tanh and sigmoid activation functions.  Because ReLU is unbounded as $x \rightarrow \infty$, it is less susceptible to vanishing gradients than tanh and sigmoid are.  It is also simple to calculate, which leads to faster training times.  

Activation function design continues to be an active area of research, and a number of novel activation functions have been introduced since ReLU, each with different properties \citep{nwankpa2018activation}.  In certain settings, these novel activation functions lead to substantial improvements in accuracy over ReLU, but the gains are often inconsistent across tasks.  Because of this inconsistency, ReLU is still the most commonly used: it is reliable, even though it may be suboptimal.

The improvements and inconsistencies are due to a gradually evolving understanding of what makes an activation function effective.  For example, Leaky ReLU \citep{maas2013rectifier} allows a small amount of gradient information to flow when the input is negative.  It was introduced to prevent ReLU from creating dead neurons, i.e.\ those that are stuck at always outputting zero.  On the other hand, the ELU activation function \citep{elu} contains a negative saturation regime to control the forward propagated variance.  These two very different activation functions have seemingly contradicting properties, yet each has proven more effective than ReLU in various tasks.

There are also often complex interactions between an activation function and other neural network design choices, adding to the difficulty of selecting an appropriate activation function for a given task.  For example, \citet{DBLP:conf/iclr/RamachandranZL18} warned that the scale parameter in batch normalization \citep{ioffe2015batch} should be set when training with the Swish activation function; \citet{hendrycks2016gaussian} suggested using an optimizer with momentum when using GELU; \citet{selu} introduced a modification of dropout \citep{hinton2012improving} called alpha dropout to be used with SELU. These results suggest that significant gains are possible by designing the activation function properly for a network and task, but that it is difficult to do so manually.

This paper presents an approach to automatic activation function design.  The approach is inspired by genetic programming \citep{koza1992genetic}, which describes techniques for evolving computer programs to solve a particular task.  In contrast with previous studies \citep{bingham2020gecco,DBLP:conf/iclr/RamachandranZL18,liu2020evolving,basirat2018quest}, this paper focuses on automatically discovering activation functions that are parametric.  Evolution discovers the general form of the function, while gradient descent optimizes the parameters of the function during training.  The approach, called PANGAEA (Parametric ActivatioN functions Generated Automatically by an Evolutionary Algorithm), discovers general activation functions that improve performance overall over previously proposed functions.  It also produces specialized functions for different architectures, such as Wide ResNet, ResNet, and Preactivation ResNet, that perform even better than the general functions, demonstrating its ability to customize activation functions to architectures.

\section{Related Work}

Prior work in automatic activation function discovery includes approaches based on either reinforcement learning (RL), evolutionary computation, or gradient descent.  In contrast, PANGAEA combines evolutionary computation and gradient descent into a single optimization process.  PANGAEA achieves better performance than previous work, and therefore is a promising approach.

\subsection{Reinforcement Learning} \citet{DBLP:conf/iclr/RamachandranZL18} used RL to design novel activation functions.  They discovered multiple functions, but analyzed just one in depth: $\textrm{Swish}(x) = x \cdot \sigma(x)$.  Of the top eight functions discovered, only Swish and $\max\{x, \sigma(x)\}$ consistently outperformed ReLU across multiple tasks, suggesting that improvements are possible but often task specific.

\subsection{Evolutionary Computation} \citet{bingham2020gecco} used evolution to discover novel activation functions.  Whereas their functions had a fixed graph structure, PANGAEA utilizes a flexible search space that implements activation functions as arbitrary computation graphs.  PANGAEA also includes more powerful mutation operations, and a function parameterization approach that makes it possible to further refine functions through gradient descent.

\citet{liu2020evolving} evolved normalization-activation layers.  They searched for a computation graph that replaced both batch normalization and ReLU in multiple neural networks.  They argued that the inherent nonlinearity of the discovered layers precluded the need for any explicit activation function.  However, experiments in this paper show that carefully designed parametric activation functions can in fact be a powerful augmentation to existing deep learning models.

\citet{basirat2018quest} used a genetic algorithm to discover task-specific piecewise activation functions.  They showed that different functions are optimal for different tasks.  However, the discovered activation functions did not outperform ELiSH and HardELiSH, two hand-designed activation functions proposed in the same paper \citep{basirat2018quest}.  The larger search space in PANGAEA affords evolution extra flexibility in designing activation functions, while the trainable parameters give customizability to the network itself, leading to consistent, significant improvement.

\subsection{Gradient Descent} Learnable activation functions (LAFs) encode functions with general functional forms such as polynomial, rational, or piecewise linear, and utilize gradient descent to discover optimal parameterizations during training \citep{apl-agostinelli2014learning, pade-molina2019pad, goyal1906learning, tavakoli2020splash}.  The general forms allow most LAFs to approximate arbitrary continuous functions.  In theory, a LAF could approximate any activation function discovered with PANGAEA.  However, just because a LAF can represent an activation function does not guarantee that the optimal activation function will be discovered by gradient descent.  Indeed, by synergizing evolutionary search and gradient descent, PANGAEA is able to outperform existing LAF approaches.

\section{The PANGAEA Method}

Activation functions in PANGAEA are represented as computation graphs, which allow for comprehensive search, efficient implementation, and effective parameterization. Regularized evolution with reranking is used as the search method to encourage exploration and to reduce noise.

\subsection{Representing and Modifying Activation Functions}

Activation functions are represented as computation graphs in which each node is a unary or a binary operator (Table \ref{tab:searchspace}).  All of these operators have TensorFlow \cite{abadi2016tensorflow} implementations, which allows for taking advantage of under-the-hood optimizations.  Safe operator implementations are chosen when possible (e.g.\ the binary operator $x_1 / x_2$ is implemented as \texttt{tf.math.divide\_no\_nan}, which returns $0$ if $x_2 = 0$).  Operators that are periodic (e.g.\ $\sin(x)$) and operators that contain repeated asymptotes are not included; in preliminary experiments they often caused training instability.  All of the operators have domain $\mathbb{R}$, making it possible to compose them arbitrarily.  The operators in Table \ref{tab:searchspace} were chosen to create a large and expressive search space that contains activation functions unlikely to be discovered by hand.  Indeed, all piecewise real analytic functions can be represented with a PANGAEA computation graph (Theorem \ref{thm:piecewise_real_analytic} of \ref{ap:proofs}).

\begin{table*}
    \centering
    \caption{The operator search space consists of basic unary and binary functions as well as existing activation functions (\ref{ap:baseline}).  $\sigma(x) = (1+e^{-x})^{-1}$.
    The unary operators bessel\_i0e and bessel\_i1e are the exponentially scaled modified Bessel functions of order 0 and 1, respectively.\newline}
    \small
    \begin{adjustbox}{max width=\textwidth}
    \begin{tabular}{lllllllll} \toprule 
        \multicolumn{7}{c}{\textbf{Unary}} & \multicolumn{2}{c}{\textbf{Binary}} \\ \midrule
        $0$           & $|x|$    & $\textrm{erf}(x)$  & $\textrm{tanh}(x)$ & $\textrm{arcsinh}(x)$ & $\textrm{ReLU}(x)$  & $\textrm{Softplus}(x)$    & $x_1 + x_2$ & $x_1^{x_2}$ \\
        $1$           & $x^{-1}$ & $\textrm{erfc}(x)$ & $e^x-1$ & $\textrm{arctan}(x)$ & $\textrm{ELU}(x)$   & $\textrm{Softsign}(x)$    & $x_1 - x_2$ & $\max\{x_1, x_2\}$ \\
        $x$           & $x^2$    & $\textrm{sinh}(x)$ & $\sigma(x)$ & $\textrm{bessel\_i0e}(x)$ & $\textrm{SELU}(x)$  & $\textrm{HardSigmoid}(x)$ & $x_1 \cdot x_2$ & $\min\{x_1, x_2\}$\\
        $-x$          & $e^x$    & $\textrm{cosh}(x)$ & $\log(\sigma(x))$ & $\textrm{bessel\_i1e}(x)$ & $\textrm{Swish}(x)$ &                           & $x_1 / x_2$ & \\
        \bottomrule
    \end{tabular}
    \end{adjustbox}
    \label{tab:searchspace}
\end{table*}

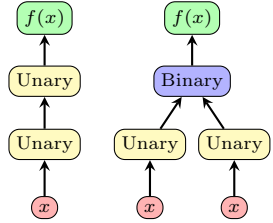
\begin{figure}
    \centering
    {\scriptsize
    \begin{tikzpicture}[node distance=3em]
    
        \node (output2) [output] {$f(x)$};
        \node (unary4) [unary, below of=output2] {Unary};
        \node (unary3) [unary, below of=unary4] {Unary};
        \node (input3) [input, below of=unary3] {$x$};

        \draw [arrow] (input3) -- (unary3);
        \draw [arrow] (unary3) -- (unary4);
        \draw [arrow] (unary4) -- (output2);

        \node (output) [output, right of=output2, xshift=4em] {$f(x)$};
        \node (binary) [binary, below of=output] {Binary};
        \node (unary1) [unary, below of=binary, xshift=-2em] {Unary};
        \node (unary2) [unary, below of=binary, xshift=2em] {Unary};
        \node (input1) [input, below of=unary1] {$x$};
        \node (input2) [input, below of=unary2] {$x$};
        
        \draw [arrow] (input1) -- (unary1);
        \draw [arrow] (input2) -- (unary2);
        \draw [arrow] (unary1) -- (binary);
        \draw [arrow] (unary2) -- (binary);
        \draw [arrow] (binary) -- (output);

    \end{tikzpicture}
    }
    \caption{Random activation function initialization. The initial population consists of random samples of two kinds of computation graphs, randomly initialized with the operators in Table~\ref{tab:searchspace}. In this manner, the search starts with simple graphs and gradually expands to more complex forms.
    }
    
    \label{fig:initialization}
\end{figure}

PANGAEA begins with an initial population of $P$ random activation functions.  Each function is either of the form $f(x) = \texttt{unary1}(\texttt{unary2}(x))$ or $f(x) = \texttt{binary}(\texttt{unary1}(x), \texttt{unary2}(x))$, as shown in Figure~\ref{fig:initialization}.  Both forms are equally likely, and the unary and binary operators are also selected uniformly at random.  The computation graphs in Figure~\ref{fig:initialization} represent the simplest non-trivial computation graphs with and without a binary operator.  This design choice is inspired by previous work in neuroevolution, which demonstrated the power of starting from simple structures and gradually complexifying them \cite{stanley2002evolving}.

During the search, all ReLU activation functions in a given neural network are replaced with a candidate activation function.  No other changes to the network or training setup are made.  The network is trained on the dataset, and the activation function is assigned a fitness score equal to the network's accuracy on the validation set.

Given a parent activation function, a child activation function is created by applying one of four possible mutations (Figure~\ref{fig:mutation}).  The possible mutations include elementary graph modifications like inserting, removing, or changing a node.  These mutations are useful for local exploration.  A special ``regenerate'' mutation is also introduced to accelerate exploration.  Other possible evolutionary operators like crossover are not used in this paper.  All mutations are equally likely with two special cases.  If a remove mutation is selected for an activation function with just one node, a change mutation is applied instead.  Additionally, if an activation function with greater than seven nodes is selected, the mutation is a remove mutation, in order to reduce bloat.

\subsubsection{Insert} In an insert mutation, one operator in the search space is selected uniformly at random.  This operator is placed on a random edge of a parent activation function graph.  In Figure~\ref{fig:mutation}\emph{b}, the unary operator $\textrm{Swish}(x)$ is inserted at the edge connecting the output of $\tanh(x)$ to the input of $x_1 + x_2$.  After mutating, the parent activation function $(\tanh(x) + |\textrm{erf}(x)|)^2$ produces the child activation function $(\textrm{Swish}(\tanh(x)) + |\textrm{erf}(x)|)^2$.  If a binary operator is randomly chosen for the insertion, the incoming input value is assigned to the variable $x_1$.  If the operator is addition or subtraction, the input to $x_2$ is set to $0$.  If the operator is multiplication, division, or exponentiation, the input to $x_2$ is set to $1$.  Finally, if the operator is the maximum or minimum operator, the input to $x_2$ is a copy of the input to $x_1$.  When a binary operator is inserted into a computation graph, the activation function computed remains unchanged.  However, the structure of the computation graph is modified and can be further altered by future mutations.

\subsubsection{Remove} 
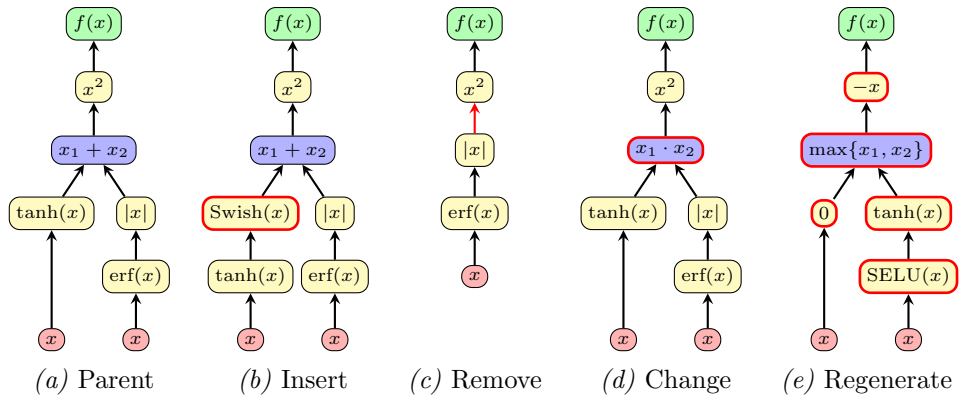
\begin{figure}
    \centering
    \scriptsize
    \begin{adjustbox}{max width=\linewidth}
    \begin{tabular}{ccccc}
    \begin{tikzpicture}[node distance=3em]
    
        \node [output] (out1) {$f(x)$};
        \node [unary, below of=out1] (u1) {$x^2$};
        \node [binary, below of=u1] (b1) {$x_1 + x_2$};
        \node [unary, below of=b1, xshift=-2em] (u2) {$\textrm{tanh}(x)$};
        \node [unary, below of=b1, xshift=2em] (u3) {$|x|$};
        \node [unary, below of=u3] (u4) {$\textrm{erf}(x)$};
        \node [input, below of=u4] (i1) {$x$};
        \node [input, left of=i1, xshift=-1em] (i2) {$x$};
        
        \node [invisible, below of=out1, yshift=-14em] (d) {\normalsize \emph{(a)} Parent};
        
        \draw [arrow] (i1) -- (u4);
        \draw [arrow] (i2) -- (u2);
        \draw [arrow] (u4) -- (u3);
        \draw [arrow] (u3) -- (b1);
        \draw [arrow] (u2) -- (b1);
        \draw [arrow] (b1) -- (u1);
        \draw [arrow] (u1) -- (out1);
    
    \end{tikzpicture}
    
    &
    
    \begin{tikzpicture}[node distance=3em]
    
        \node [output] (out1) {$f(x)$};
        \node [unary, below of=out1] (u1) {$x^2$};
        \node [binary, below of=u1] (b1) {$x_1 + x_2$};
        \node [unary, below of=b1, xshift=-2em, draw=red, line width=1] (swish) {$\textrm{Swish}(x)$};
        \node [unary, below of=swish] (u2) {$\textrm{tanh}(x)$};
        \node [unary, below of=b1, xshift=2em] (u3) {$|x|$};
        \node [unary, below of=u3] (u4) {$\textrm{erf}(x)$};
        \node [input, below of=u4] (i1) {$x$};
        \node [input, left of=i1, xshift=-1em] (i2) {$x$};
        
        \node [invisible, below of=out1, yshift=-14em] (d) {\normalsize \emph{(b)} Insert};
        
        \draw [arrow] (i1) -- (u4);
        \draw [arrow] (i2) -- (u2);
        \draw [arrow] (u4) -- (u3);
        \draw [arrow] (u3) -- (b1);
        \draw [arrow] (u2) -- (swish);
        \draw [arrow] (swish) -- (b1);
        \draw [arrow] (b1) -- (u1);
        \draw [arrow] (u1) -- (out1);
    
    \end{tikzpicture}
    
    &
    
    \begin{tikzpicture}[node distance=3em]
    
        \node [output] (out1) {$f(x)$};
        \node [unary, below of=out1] (u1) {$x^2$};
        \node [unary, below of=u1] (u3) {$|x|$};
        \node [unary, below of=u3] (u4) {$\textrm{erf}(x)$};
        \node [input, below of=u4] (i1) {$x$};
        
        \node [invisible, below of=out1, yshift=-14em] (d) {\normalsize \emph{(c)} Remove};
        
        \draw [arrow] (i1) -- (u4);
        \draw [arrow] (u4) -- (u3);
        \draw [arrow, draw=red] (u3) -- (u1);
        \draw [arrow] (u1) -- (out1);
    
    \end{tikzpicture}
    
    &
    
    \begin{tikzpicture}[node distance=3em]
    
        \node [output] (out1) {$f(x)$};
        \node [unary, below of=out1] (u1) {$x^2$};
        \node [binary, below of=u1, draw=red, line width=1] (b1) {$x_1 \cdot x_2$};
        \node [unary, below of=b1, xshift=-2em] (u2) {$\textrm{tanh}(x)$};
        \node [unary, below of=b1, xshift=2em] (u3) {$|x|$};
        \node [unary, below of=u3] (u4) {$\textrm{erf}(x)$};
        \node [input, below of=u4] (i1) {$x$};
        \node [input, left of=i1, xshift=-1em] (i2) {$x$};
        
        \node [invisible, below of=out1, yshift=-14em] (d) {\normalsize \emph{(d)} Change};
        
        \draw [arrow] (i1) -- (u4);
        \draw [arrow] (i2) -- (u2);
        \draw [arrow] (u4) -- (u3);
        \draw [arrow] (u3) -- (b1);
        \draw [arrow] (u2) -- (b1);
        \draw [arrow] (b1) -- (u1);
        \draw [arrow] (u1) -- (out1);
    
    \end{tikzpicture}
    
    &

    \begin{tikzpicture}[node distance=3em]
    
        \node [output] (out1) {$f(x)$};
        \node [unary, below of=out1, draw=red, line width=1] (u1) {$-x$};
        \node [binary, below of=u1, draw=red, line width=1] (b1) {$\max\{x_1, x_2\}$};
        \node [unary, below of=b1, xshift=-2em, draw=red, line width=1] (u2) {$0$};
        \node [unary, below of=b1, xshift=2em, draw=red, line width=1] (u3) {$\textrm{tanh}(x)$};
        \node [unary, below of=u3, draw=red, line width=1] (u4) {$\textrm{SELU}(x)$};
        \node [input, below of=u4] (i1) {$x$};
        \node [input, left of=i1, xshift=-1em] (i2) {$x$};
        
        \node [invisible, below of=out1, yshift=-14em] (d) {\normalsize \emph{(e)} Regenerate};
        
        \draw [arrow] (i1) -- (u4);
        \draw [arrow] (i2) -- (u2);
        \draw [arrow] (u4) -- (u3);
        \draw [arrow] (u3) -- (b1);
        \draw [arrow] (u2) -- (b1);
        \draw [arrow] (b1) -- (u1);
        \draw [arrow] (u1) -- (out1);
    
    \end{tikzpicture}

    \end{tabular}
    \end{adjustbox}
    \caption{Evolutionary operations on activation functions. In an `Insert' mutation, a new operator is inserted in one of the edges of the computation graph, like the Swish$(x)$ in \emph{(b)}. In a `Remove' mutation, a node in the computation graph is deleted, like the addition in \emph{(c)}. In a `Change' mutation, an operator at a node is replaced with another, like addition with multiplication in \emph{(d)}. These first three mutations are useful in refining the function locally. In contrast, in a `Regenerate' mutation \emph{(e)}, every operator in the graph is replaced by a random operator, thus increasing exploration.}
    \label{fig:mutation}
\end{figure}

In a remove mutation, one node is selected uniformly at random and deleted.  The node's input is rewired to its output.  If the removed node is binary, one of the two inputs is chosen at random and is deleted.  The other input is kept.  In Figure~\ref{fig:mutation}\emph{c}, the addition operator is removed from the parent activation function.  The two inputs to addition, $\tanh(x)$ and $|\textrm{erf}(x)|$, cannot both be kept.  By chance, $\tanh(x)$ is discarded, resulting in the child activation function $|\textrm{erf}(x)|^2$.  

\subsubsection{Change} To perform a change mutation, one node in the computation graph is selected at random and replaced with another operator from the search space, also uniformly at random.  Unary operators are always replaced with unary operators, and binary operators with binary operators.  Figure~\ref{fig:mutation}\emph{d} shows how changing addition to multiplication produces the activation function $(\tanh(x) \cdot |\textrm{erf}(x)|)^2$.

\subsubsection{Regenerate} In a regenerate mutation, every operator in the computation graph is replaced with another operator from the search space.  As with change mutations, unary operators are replaced with unary operators, and binary operators with binary operators.  Although every node in the graph is changed, the overall structure of the computation graph remains the same.  Regenerate mutations are useful for increasing exploration, and are similar in principle to burst mutation and delta coding \citep{gomezgecco03,whitleyicga91}.  Figure~\ref{fig:mutation}\emph{e} shows the child activation function $-\max\{0, \tanh(\textrm{SELU}(x))\}$, which is quite different from the parent function in Figure~\ref{fig:mutation}\emph{a}.

\subsubsection{Parameterization of Activation Functions}

\begin{figure}
    \centering
    \begin{multicols}{2}
    \null \vfill 
    {\scriptsize
     \begin{tikzpicture}[node distance=3em]
    
        \node (output) [output] {$f(x)$};
        \node (unary) [unary, below of=output] {$\sigma(x)$};
        \node (binary) [binary, below of=unary] {$x_1 - x_2$};
        \node (unary1) [unary, below of=binary, xshift=-2em] {$|x|$};
        \node (unary2) [unary, below of=binary, xshift=2em] {$\textrm{arctan}(x)$};
        \node (input1) [input, below of=unary1] {$x$};
        \node (input2) [input, below of=unary2] {$x$};

        \draw [arrow] (input1) -- (unary1);
        \draw [arrow, draw=red] (input2) -- (unary2);
        \draw [arrow, draw=red] (unary1) -- (binary);
        \draw [arrow] (unary2) -- (binary);
        \draw [arrow] (binary) -- (unary);
        \draw [arrow, draw=red] (unary) -- (output);
        
    \end{tikzpicture}}
    
    \vfill \null
    {\scriptsize
    \begin{tikzpicture}[node distance=3em]
    
        \node (output) [output] {$f(x)$};
        \node (p1) [maybeparam, below of=output] {$\alpha$};
        \node (unary) [unary, below of=p1] {$\sigma(x)$};
        \node (binary) [binary, below of=unary] {$x_1 - x_2$};
        \node (p5) [maybeparam, below of=binary, xshift=-2em] {$\beta$};
        \node (unary1) [unary, below of=p5] {$|x|$};
        \node (unary2) [unary, below of=binary, xshift=2em] {$\textrm{arctan}(x)$};
        \node (p4) [maybeparam, below of=unary2] {$\gamma$};
        \node (input1) [input, below of=unary1] {$x$};
        \node (input2) [input, below of=p4] {$x$};

        \draw [arrow] (input1) -- (unary1);
        \draw [arrow] (input2) -- (p4);
        \draw [arrow] (p4) -- (unary2); 
        \draw [arrow] (unary1) -- (p5);
        \draw [arrow] (p5) -- (binary);
        \draw [arrow] (unary2) -- (binary);
        \draw [arrow] (binary) -- (unary);
        \draw [arrow] (unary) -- (p1);
        \draw [arrow] (p1) -- (output);
        
    \end{tikzpicture}
    }
    
    \end{multicols}
    \caption{Parameterization of activation functions.  In this example, parameters are added to $k=3$ random edges, yielding the parametric activation function $\alpha \sigma(\beta |x| - \textrm{arctan}(\gamma x))$.}
    \label{fig:parameterization}
\end{figure}
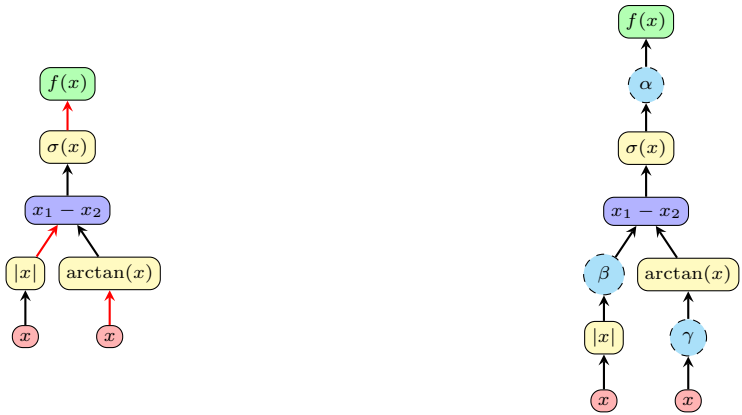

After mutation (or random initialization), activation functions are
parameterized (Figure~\ref{fig:parameterization}).  A value
$k \in \{0, 1, 2, 3\}$ is chosen uniformly at random, and $k$ edges of
the activation function graph are randomly selected.
Multiplicative per-channel parameters are inserted at these edges and
initialized to one.  Whereas evolution is well suited for discovering the
general form of the activation function in a discrete, structured
search space, parameterization makes it possible to fine-tune the
function using gradient descent.
The function parameters are updated at	every epoch during backpropagation, resulting in
different activation functions in different stages of training.  As the parameters are per-channel, the process
creates different activation functions at different locations in the
neural network.  Thus, parameterization gives neural networks additional flexibility to customize activation functions.

\subsection{Discovering Activation Functions with Evolution}
\label{sec:evolution}

Activation functions are discovered by regularized evolution \citep{real2019regularized}.  Initially, $P$ random activation functions are created, parameterized, and assigned fitness scores.
To generate a new activation function, $S$ functions are sampled with
replacement from the current population.  The function with the
highest validation accuracy serves as the parent, and is mutated to create a child activation
function.  This function is parameterized and assigned a fitness
score.  The new activation function is then added to the population, and the
oldest function in the population is removed, ensuring the
population is always of size $P$.  This process continues until $C$ functions have been
evaluated in total, and the top functions over the history
of the search are returned as a result.  

Any activation function that
achieves a fitness score less than a threshold $V$ is discarded.
These functions are not added to the population, but they
do count towards the total number of $C$ activation functions
evaluated for each architecture.  This quality control mechanism
allows evolution to focus only on the most promising candidates.

To save computational resources during evolution, each activation function is evaluated by training a neural network for 100 epochs using a compressed learning rate schedule. After evolution is complete, the top 10 activation functions from the entire search are reranked.  Each function receives an adjusted fitness score equal to the average validation accuracy from two independent 200-epoch training runs using the original learning rate schedule.  The top three activation functions after reranking proceed to the final testing experiments.

During evolution, it is possible that some activation functions achieve unusually high validation accuracy by chance.  The 100-epoch compressed learning rate schedule may also have a minor effect on which activation functions are optimal compared to a full 200-epoch schedule.  Reranking thus serves two purposes.  Full training reduces bias from the compressed schedule, and averaging two such runs lessens the impact of activation functions that achieved high accuracy by chance.

\section{Datasets and Architectures}

The experiments in this paper focus primarily on the CIFAR-100 image classification dataset \citep{krizhevsky2009learning}. This dataset is a more difficult version of the popular CIFAR-10 dataset, with 100 object categories instead of 10. Fifty images from each class were randomly selected from the training set to create a balanced validation set, resulting in a training/validation/test split of 45K/5K/10K images.

To demonstrate that PANGAEA can discover effective activation functions in various settings, it is evaluated with three different neural networks.  The models were implemented in TensorFlow \citep{abadi2016tensorflow}, mirroring the original authors' training setup as closely as possible (see \ref{ap:details} for training details and \ref{ap:custom} for code that shows how to train with custom activation functions).

{\bf Wide Residual Network}
\citep[WRN-10-4;][]{zagoruyko2016wide} has a depth of 10 and widening factor of four.  Wide residual networks provide an interesting comparison because they are shallower and wider than many other popular architectures, while still achieving good results.  WRN-10-4 was chosen because its CIFAR-100 accuracy is competitive, yet it trains relatively quickly.

{\bf Residual Network}
\citep[ResNet-v1-56;][]{he2016deep}, with a depth of 56, provides an important contrast to WRN-10-4.  It is significantly deeper and has a slightly different training setup, which may have an effect on the performance of different activation functions.  

{\bf Preactivation Residual Network}
\citep[ResNet-v2-56;][]{he2016identity} has identical depth to ResNet-v1-56, but is a fundamentally different architecture.  Activation functions are not part of the skip connections, as is the case in ResNet-v1-56.  Since information does not have to pass through an activation function, this structure makes it easier to train very deep architectures.  PANGAEA should exploit this structure and discover different activation functions for ResNet-v2-56 and ResNet-v1-56.

\section{Main Results}
\label{sec:results}

\subsection{Overview}

Separate evolution experiments were run to discover novel
activation functions for each of the three architectures. Evolutionary
parameters $P=64$, $S=16$, $C=1{,}000$, and $V=20\%$ were used since
they were found to work well in preliminary experiments.

\begin{figure}
    \centering
    \includegraphics[width=\linewidth]{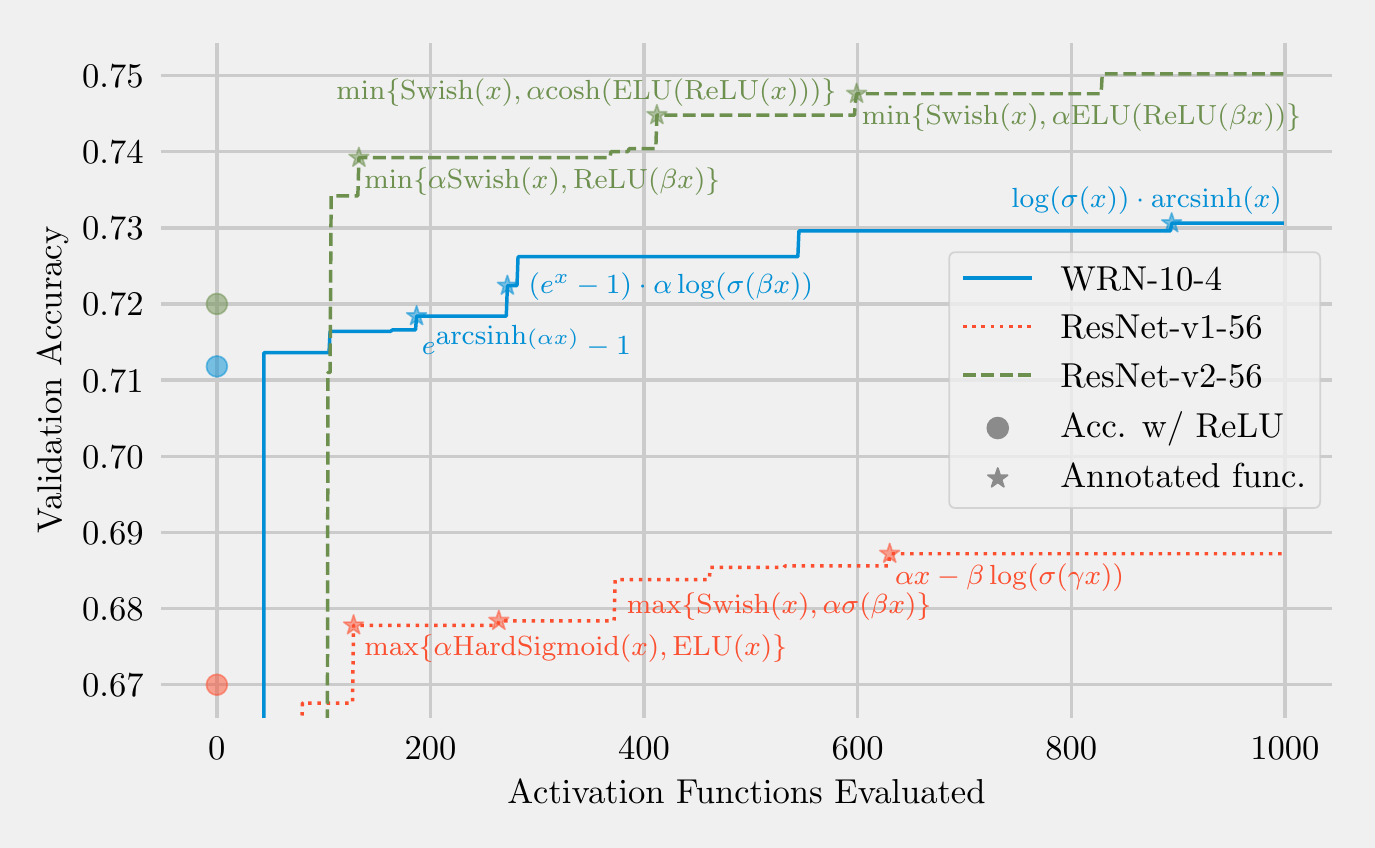}
    \caption{Progress of PANGAEA with three different neural networks.  The plots show the best accuracy achieved among all activation functions evaluated so far.  The stars on the plot specify the time when notable activation functions were discovered during evolution; the expression for each such function is written next to the star. Evolution quickly discovered activation functions that outperform      
      ReLU (accuracy with ReLU shown at $x=0$), 
      and continued to improve throughout the experiment.   Note that this figure shows validation accuracy, while Table~\ref{tab:results} lists test set accuracy.}
    \label{fig:evolution}
\end{figure}

Figure~\ref{fig:evolution} visualizes progress in these experiments.
For all three architectures, PANGAEA quickly discovered
activation functions that outperform ReLU.  It continued to make
further progress, gradually discovering better activation functions, and did not
plateau during the time allotted for the experiment.  Each run
took approximately 1,000 GPU hours on GeForce GTX 1080 and 1080 Ti GPUs
(see \ref{ap:implementation} for implementation and compute details).

Table \ref{tab:results} shows the final test accuracy for the top specialized activation functions discovered by PANGAEA in each run. For comparison, the accuracy of the top general functions discovered in this process are also shown, as well as the accuracy of several baseline activation functions (see \ref{ap:baseline} for baseline activation function details and \ref{ap:pausplash} for additional results with learnable baseline functions). In sum, PANGAEA discovered the best activation function for each of the three architectures.

\subsection{Specialized Activation Functions}
For all three architectures, there are baseline activation functions that outperform ReLU by statistically significant margins.  This result already demonstrates that activation functions should be chosen carefully, and that the common practice of using ReLU by default is suboptimal.  Furthermore, the best baseline activation function is different for different architectures, suggesting that specializing activation functions to the architecture is a good approach.

Because PANGAEA uses validation accuracy from a single neural network to assign fitness scores to activation functions, there is selective pressure to discover functions that exploit the structure of the network.  The functions thus become specialized to the architecture. They increase the performance of that architecture; however, they may not be as effective with other architectures.  Specialized activation function accuracies are highlighted with the gray background in Table \ref{tab:results}.  To verify that the functions are customized to a specific architecture, the functions were cross-evaluated with other architectures.  

\begin{table*}[!ht]
    \centering
    \renewcommand{\arraystretch}{.6666667} 
    \caption{CIFAR-100 test set accuracy aggregated over ten runs, shown as mean $\pm$ sample standard deviation.  Asterisks indicate a statistically significant improvement in mean accuracy over ReLU, with * if $p \leq 0.05$, ** if $p \leq 0.01$, and *** if $p \leq 0.001$;  $p$-values are from one-tailed Welch's $t$-tests.  The top accuracy for each architecture is in bold.  Baseline activation function details and references are given in \ref{ap:baseline}.\newline
    }
    \centering
    \begin{adjustbox}{max width=\textwidth}
    \small
    \begin{tabular}{llll} \toprule
    & \textbf{WRN-10-4} & \textbf{ResNet-v1-56} & \textbf{ResNet-v2-56} \\ \midrule

    \textbf{Specialized for WRN-10-4}\\
    $\log(\sigma(\alpha x)) \cdot \beta \textrm{arcsinh}(x)$ & 
    \cellcolor{black!5} $\bm{73.20} {\scriptscriptstyle \pm 0.37~***}$ &
    $18.63 {\scriptscriptstyle \pm 21.04}$ &
    $45.88 {\scriptscriptstyle \pm 30.70}$ \\
    
    $\log(\sigma(\alpha x)) \cdot \textrm{arcsinh}(x)$ & 
    \cellcolor{black!5} $73.16 {\scriptscriptstyle \pm 0.41~***}$ &
    $19.34 {\scriptscriptstyle \pm 20.14}$ &
    $64.30 {\scriptscriptstyle \pm 21.32}$ \\
    
    $-\textrm{Swish}(\textrm{Swish}(\alpha x))$ & 
    \cellcolor{black!5} $72.49 {\scriptscriptstyle \pm 0.55~***}$ &
    $58.86 {\scriptscriptstyle \pm 2.88}$ &
    $74.71 {\scriptscriptstyle \pm 0.20~*}$ \\ 
    \midrule 

    \textbf{Specialized for ResNet-v1-56}\\
    $\alpha x - \beta \log(\sigma(\gamma x))$ & 
    $70.28 {\scriptscriptstyle \pm 0.37}$ &
    \cellcolor{black!5} $\bm{71.01} {\scriptscriptstyle \pm 0.64~***}$ &
    $74.35 {\scriptscriptstyle \pm 0.45}$ \\
    
    $\alpha x - \log(\sigma(\beta x))$ & 
    $70.47 {\scriptscriptstyle \pm 0.53}$ &
    \cellcolor{black!5} $70.30 {\scriptscriptstyle \pm 0.58~*}$ &
    $74.70 {\scriptscriptstyle \pm 0.23~*}$ \\
    
    $\max\{\textrm{Swish}(x), 0\}$ & 
    $72.10 {\scriptscriptstyle \pm 0.33~**}$ &
    \cellcolor{black!5} $69.43 {\scriptscriptstyle \pm 0.69}$ &
    $74.97 {\scriptscriptstyle \pm 0.25~**}$ \\
    \midrule 
    
    \textbf{Specialized for ResNet-v2-56}\\
    $\textrm{Softplus}(\textrm{ELU}(x))$ & 
    $71.36 {\scriptscriptstyle \pm 0.34}$ &
    $69.96 {\scriptscriptstyle \pm 0.39}$ &
    \cellcolor{black!5} $\bm{75.61} {\scriptscriptstyle \pm 0.42~***}$ \\
    
    $\min\{\log(\sigma(x)), \alpha \log(\sigma(\beta x))\}$ & 
    $72.04 {\scriptscriptstyle \pm 0.34~**}$ &
    $69.56 {\scriptscriptstyle \pm 0.48}$ &
    \cellcolor{black!5} $75.19 {\scriptscriptstyle \pm 0.39~***}$ \\
    
    $\textrm{SELU}(\textrm{Swish}(x))$ & 
    $01.00 {\scriptscriptstyle \pm 0.00}$ &
    $01.00 {\scriptscriptstyle \pm 0.00}$ &
    \cellcolor{black!5} $75.02 {\scriptscriptstyle \pm 0.35~**}$ \\
    \midrule 
    
    \textbf{General Activation Functions}\\
    $\max\{\textrm{Swish}(x), \alpha \log (\sigma (\textrm{ReLU}(x)))\}$ & 
     $72.54 {\scriptscriptstyle \pm 0.26~***}$ &
     $69.91 {\scriptscriptstyle \pm 0.37}$ &
     $75.20 {\scriptscriptstyle \pm 0.41~***}$ \\
    
    $\min\{\textrm{Swish}(x), \alpha \textrm{ELU}(\textrm{ReLU}(\beta x))\}$ & 
     $72.39 {\scriptscriptstyle \pm 0.29~***}$ &
     $69.82 {\scriptscriptstyle \pm 0.40}$ &
     $75.27 {\scriptscriptstyle \pm 0.38~***}$ \\
    
    $\log(\sigma(x))$ & 
     $72.33 {\scriptscriptstyle \pm 0.32~***}$ &
     $69.58 {\scriptscriptstyle \pm 0.35}$ &
     $75.53 {\scriptscriptstyle \pm 0.37~***}$ \\
    \midrule
    
    \textbf{Fixed Baseline Functions}\\
    $\textrm{ReLU}$ &
    $71.46 {\scriptscriptstyle \pm 0.50}$ &
    $69.64 {\scriptscriptstyle \pm 0.65}$ &
    $74.39 {\scriptscriptstyle \pm 0.44}$ \\

    $\textrm{ELiSH}$ & 
    $01.00 {\scriptscriptstyle \pm 0.00}$ &
    $01.00 {\scriptscriptstyle \pm 0.00}$ &
    $75.20 {\scriptscriptstyle \pm 0.31~***}$ \\
    
    $\textrm{ELU}$ & 
    $72.30 {\scriptscriptstyle \pm 0.32~***}$ &
    $69.67 {\scriptscriptstyle \pm 0.46}$ &
    $74.95 {\scriptscriptstyle \pm 0.30~**}$ \\
    
    $\textrm{GELU}$ & 
    $71.95 {\scriptscriptstyle \pm 0.35~*}$ &
    $70.19 {\scriptscriptstyle \pm 0.40~*}$ &
    $74.86 {\scriptscriptstyle \pm 0.33~**}$ \\
    
    $\textrm{HardSigmoid}$ & 
    $54.99 {\scriptscriptstyle \pm 1.00}$ &
    $32.55 {\scriptscriptstyle \pm 4.06}$ &
    $64.90 {\scriptscriptstyle \pm 0.69}$ \\
    
    $\textrm{Leaky ReLU}$ & 
    $71.73 {\scriptscriptstyle \pm 0.33}$ &
    $69.78 {\scriptscriptstyle \pm 0.33}$ &
    $74.73 {\scriptscriptstyle \pm 0.35~*}$ \\
    
    $\textrm{Mish}$ & 
    $71.95 {\scriptscriptstyle \pm 0.41~*}$ &
    $69.88 {\scriptscriptstyle \pm 0.54}$ &
    $75.32 {\scriptscriptstyle \pm 0.29~***}$ \\
    
    $\textrm{SELU}$ & 
    $70.53 {\scriptscriptstyle \pm 0.42}$ &
    $68.52 {\scriptscriptstyle \pm 0.29}$ &
    $73.79 {\scriptscriptstyle \pm 0.36}$ \\
    
    $\textrm{sigmoid}$ & 
    $56.10 {\scriptscriptstyle \pm 0.98}$ &
    $36.47 {\scriptscriptstyle \pm 3.32}$ &
    $66.45 {\scriptscriptstyle \pm 0.92}$ \\
    
    $\textrm{Softplus}$ & 
    $72.27 {\scriptscriptstyle \pm 0.26~***}$ &
    $69.71 {\scriptscriptstyle \pm 0.36}$ &
    $75.46 {\scriptscriptstyle \pm 0.52~***}$ \\
    
    $\textrm{Softsign}$ & 
    $56.30 {\scriptscriptstyle \pm 2.16}$ &
    $58.38 {\scriptscriptstyle \pm 0.96}$ &
    $69.33 {\scriptscriptstyle \pm 0.39}$ \\
    
    $\textrm{Swish}$ & 
    $72.26 {\scriptscriptstyle \pm 0.28~***}$ &
    $69.68 {\scriptscriptstyle \pm 0.38}$ &
    $75.08 {\scriptscriptstyle \pm 0.36~***}$ \\
    
    $\textrm{tanh}$ & 
    $56.52 {\scriptscriptstyle \pm 1.53}$ &
    $63.88 {\scriptscriptstyle \pm 0.38}$ &
    $70.44 {\scriptscriptstyle \pm 0.40}$ \\
    \midrule
    
    \textbf{Parametric Baseline Functions}\\
    
    $\textrm{PReLU}$ &
    $72.23 {\scriptscriptstyle \pm 0.37~***}$ &
    $69.77 {\scriptscriptstyle \pm 0.40}$ &
    $75.10 {\scriptscriptstyle \pm 0.53~**}$ \\

    $\textrm{PSwish} = x \cdot \sigma(\beta x)$ & 
    $72.40 {\scriptscriptstyle \pm 0.31~***}$ &
    $70.16 {\scriptscriptstyle \pm 0.46~*}$ &
    $75.39 {\scriptscriptstyle \pm 0.28~***}$ \\
    \midrule
    
    \textbf{Learnable Baseline Functions}\\
    
    $\textrm{APL}$ &
    $72.88 {\scriptscriptstyle \pm 0.32~***}$ &
    $70.81 {\scriptscriptstyle \pm 0.20~***}$ &
    $75.02 {\scriptscriptstyle \pm 0.28~***}$ \\

    $\textrm{PAU}$ &
    $41.46 {\scriptscriptstyle \pm 22.66}$ &
    $01.00 {\scriptscriptstyle \pm 0.00}$ &
    $02.38 {\scriptscriptstyle \pm 4.36}$ \\
    
    $\textrm{SPLASH}$ &
    $72.16 {\scriptscriptstyle \pm 0.81~*}$ &
    $01.00 {\scriptscriptstyle \pm 0.00}$ &
    $73.45 {\scriptscriptstyle \pm 0.43}$ \\
    
    \bottomrule
    \end{tabular}
    \end{adjustbox}
    \label{tab:results}
\end{table*}

PANGAEA discovered two specialized activation functions for WRN-10-4 that outperformed all baseline functions by a statistically significant margin ($p \leq 0.05$).  The top specialized function on ResNet-v1-56 also significantly outperformed all baseline functions, except APL (for which $p = 0.19$).  The top specialized activation function on ResNet-v2-56 similarly significantly outperformed all except Softplus ($p = 0.25$) and PSwish ($p = 0.09$). These results strongly demonstrate the power of customizing activation functions to architectures.  Indeed, specializing activation functions is a new dimension of activation function meta-learning not considered by previous work \cite{DBLP:conf/iclr/RamachandranZL18, basirat2018quest, bingham2020gecco}.


\subsection{General Activation Functions}
Although the best performance comes from specialization, it is also useful to discover activation functions that achieve high accuracy across multiple architectures. For instance, they could be used initially on a new architecture before spending compute on specialization. A powerful albeit computationally demanding approach would be to evolve general functions directly, by evaluating candidates on multiple architectures during evolution. However, it turns out that each specialized evolution run already generates a variety of functions, many of which are general.

To evaluate whether the PANGAEA runs discovered general functions as well, the top 10 functions from each run were combined into a pool of 30 candidate functions.  Each candidate was assigned three fitness scores equal to the average validation accuracy from two independent training runs on each of the three architectures.  Candidate functions that were Pareto-dominated, were functionally equivalent to one of the baseline activation functions, or had already been selected as a specialized activation function were discarded, leaving three Pareto-optimal general activation functions.

These functions indeed turned out to be effective as general activation functions.  All three achieved good accuracy with ResNet-v1-56 and significantly outperformed ReLU with WRN-10-4 and ResNet-v2-56.  However, specialized activation functions, i.e.\ those specifically evolved for each architecture, still give the biggest improvements. 

\subsection{Shapes of Discovered Functions}

Many of the top discovered activation functions are compositions of
multiple unary operators.  These functions do not exist in the core
unit search space of \citet{DBLP:conf/iclr/RamachandranZL18}, which
requires binary operators.  They also do not exist in the $S_1$ or
$S_2$ search spaces proposed by \citet{bingham2020gecco}, which are too shallow.  
The design of the search space is therefore as
important as the search algorithm itself.  Previous search spaces that
rely on repeated fixed building blocks only have limited
representational power. In contrast, PANGAEA utilizes a flexible
search space that can represent activation functions in an arbitrary
computation graph (see \ref{ap:searchspace} for an analysis on the size of the PANGAEA search space). 

Furthermore, while the learnable baseline functions can in principle approximate the functions discovered by PANGAEA, they do not consistently match its performance.  PANGAEA utilizes both evolutionary search and gradient descent to discover activation functions, and apparently this combination of optimization processes is more powerful than gradient descent alone.
\begin{figure}
    \centering
    \includegraphics[clip, trim=1.75em 0em 5.5em 0em, width=\linewidth]{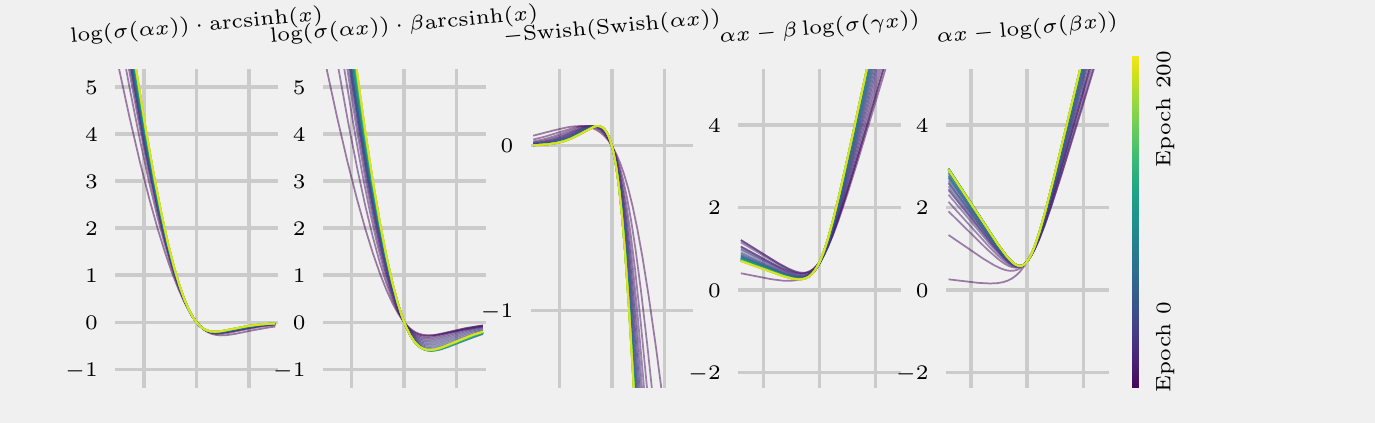}\\
    \vspace{-0.5em}
    \includegraphics[clip, trim=1.75em 0.5em 5.5em 1em, width=\linewidth]{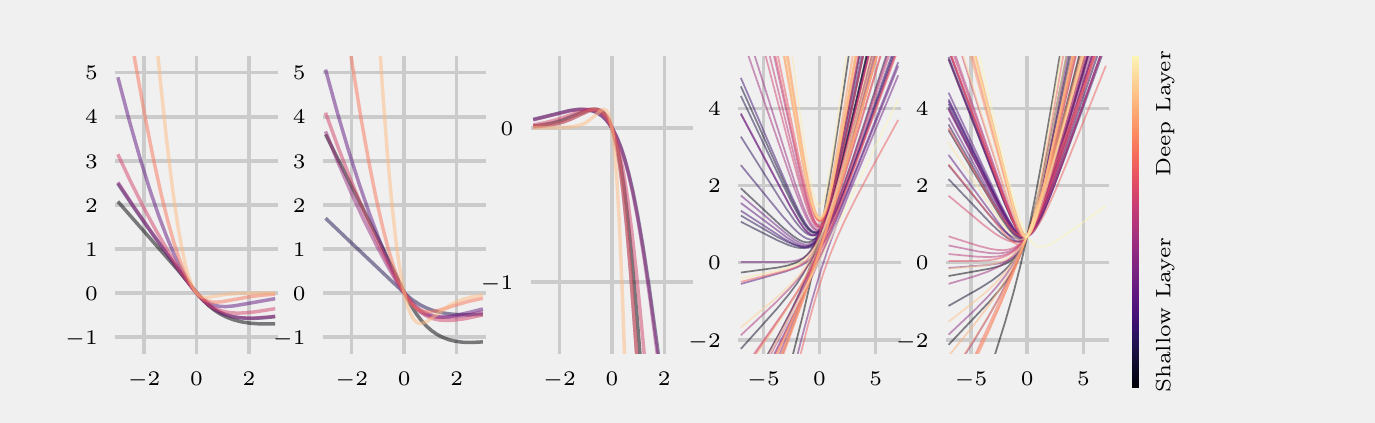}
    \vspace{-1.5em}
    \caption{Adaptation of parametric activation functions over time and        
      space. \textbf{Top:} The parameters change during training,           
      resulting in different activation functions in the early and late         
      stages. The plots were created by averaging the values of $\alpha$,       
      $\beta$, and $\gamma$ across the entire network at different training epochs. \textbf{Bottom:} The           
      parameters are updated separately in each channel, inducing               
      different activation functions at different locations of a neural         
      network. The plots were created by averaging $\alpha$, $\beta$, and       
      $\gamma$ at each layer of the network after the completion of             
      training.}
    \label{fig:varying}
\end{figure}

Figure~\ref{fig:varying} shows examples of parametric activation
functions discovered by PANGAEA.  As training progresses, gradient
descent makes small adjustments to the function parameters $\alpha$,
$\beta$, and $\gamma$, resulting in activation functions that change
over time.  This result suggests that
it is advantageous to have one activation function in the early stages
of training when the network learns rapidly, and a different
activation function in the later stages of training when the network
is focused on fine-tuning. The parameters $\alpha$, $\beta$, and
$\gamma$ are also learned separately for the different channels,
resulting in activation functions that vary with location in a neural
network. Functions in deep layers
(near the output) are more nonlinear than those in shallow
layers (closer to the input), possibly contrasting the need to
form regularized embeddings with the need to form categorizations. In
this manner, PANGAEA customizes	the activation functions to both time
and space for each architecture.

\section{Diving Deeper: Experiments on Ablations, Variations, and Training Dynamics}
\label{sec:ablations}

PANGAEA is a method with many moving parts.  The main results from Section \ref{sec:results} already showed the power of PANGAEA, and the experiments from this section illuminate how each component of PANGAEA contributes to its success.  Evolutionary search and gradient descent working in tandem provided a better strategy than either optimization algorithm alone (Section \ref{sec:baselinesearchstrategies}).  The top activation functions benefitted from their learnable parameters (Section \ref{sec:effectofparameterization}), but baseline functions did not (Section \ref{sec:parametric_baseline}), showing how PANGAEA discovered functional forms well-suited to parameterization.  PANGAEA is robust: the activation functions it discovers transfer to larger networks (Section \ref{sec:scalingup}) and PANGAEA achieves impressive results with a new dataset and architecture (Section \ref{sec:allcnnc}).  The activation functions discovered by PANGAEA improve accuracy by easing optimization and implicitly regularizing the network (Section \ref{sec:dynamics}).

\subsection{Additional Baseline Search Strategies}
\begin{table}
    \centering
    \caption{WRN-10-4 accuracy with different activation functions on CIFAR-100, shown as mean $\pm$ sample standard deviation across ten runs.  PANGAEA discovers better activation functions than random search and nonparametric evolution.\newline}
    \small
    \begin{adjustbox}{max width=\linewidth}
    \begin{tabular}{ll} \toprule
        \textbf{PANGAEA} \\
        $\log(\sigma(\alpha x)) \cdot \beta \textrm{arcsinh}(x)$ & 
        $\bm{73.20} {\scriptscriptstyle \pm 0.37}$ \\
        $\log(\sigma(\alpha x)) \cdot \textrm{arcsinh}(x)$ & 
        $73.16 {\scriptscriptstyle \pm 0.41}$ \\
        $-\textrm{Swish}(\textrm{Swish}(\alpha x))$ & 
        $72.49 {\scriptscriptstyle \pm 0.55}$ \\ 
        \midrule
        
        \textbf{Random Search}\\
        $\alpha \textrm{Swish}(x)$ & $72.85 {\scriptscriptstyle \pm 0.25}$ \\
        $\textrm{Softplus}(x) \cdot \arctan(\alpha x)$ & $72.81 {\scriptscriptstyle \pm 0.35}$ \\
        $\textrm{ReLU}(\alpha \textrm{arcsinh}(\beta \sigma(x))) \cdot \textrm{SELU}(\gamma x)$ & $72.69 {\scriptscriptstyle \pm 0.21}$ \\
        \midrule 
        
        \textbf{Nonparametric Evolution}\\
        $\cosh(1) \cdot \textrm{Swish}(x)$ & 
        $72.78 {\scriptscriptstyle \pm 0.24}$ \\    
        $(e^1-1) \cdot \textrm{Swish}(x)$ & 
        $72.52 {\scriptscriptstyle \pm 0.34}$ \\
        $\textrm{ReLU}(\textrm{Swish}(x))$ &
        $72.04 {\scriptscriptstyle \pm 0.54}$ \\
        \midrule
        
        ReLU & $71.46 {\scriptscriptstyle \pm 0.50}$ \\
        Swish & $72.26 {\scriptscriptstyle \pm 0.28}$ \\

    \bottomrule
    \end{tabular}
    \end{adjustbox}
    \label{tab:search_strategy}
\end{table}
\label{sec:baselinesearchstrategies}

As additional baseline comparisons, two alternative search strategies were used to discover activation functions for WRN-10-4.  First, a random search baseline was established by applying random mutations without regard to fitness values.  This approach corresponds to setting evolutionary parameters $P=1$, $S=1$, and  $V=0\%$.  Second, to understand the effects of function parameterization, a nonparametric evolution baseline was run.  This setting is identical to PANGAEA, except functions are not parameterized (Figure \ref{fig:parameterization}).  Otherwise, both baselines follow the same setup as PANGAEA, including evaluating $C=1{,}000$ candidate functions and reranking the most promising ones (Section \ref{sec:evolution}).

Table \ref{tab:search_strategy} shows the results of this experiment.  Random search is able to discover good functions that outperform ReLU, but the functions are not as powerful as those discovered by PANGAEA.  This result demonstrates the importance of fitness selection in evolutionary search.  The functions discovered by nonparametric evolution similarly outperform ReLU but underperform PANGAEA.  Interestingly, without parameterization, evolution is not as creative: two of the three functions discovered are merely Swish multiplied by a constant.  Random search and nonparametric evolution both discovered good functions that improved accuracy, but PANGAEA achieves the best performance by combining the advantages of fitness selection and function parameterization.

\subsection{Effect of Parameterization}
\begin{table}
    \centering
    \caption{CIFAR-100 test set accuracy aggregated over ten runs, shown as mean $\pm$ sample standard deviation.  The parametric evolved functions tend to outperform their nonparametric counterparts, demonstrating the value of parameterization.\newline}
    \small
    \begin{adjustbox}{max width=\linewidth}
    \begin{tabular}{ll} \toprule
         
    \multicolumn{2}{l}{\textbf{WRN-10-4}}\\
    $\log(\sigma(\alpha x)) \cdot \beta \textrm{arcsinh}(x)$ & 
    $\bm{73.20} {\scriptscriptstyle \pm 0.37}$ \\
    $\log(\sigma(\alpha x)) \cdot \textrm{arcsinh}(x)$ & 
    $73.16 {\scriptscriptstyle \pm 0.41}$ \\
    $\log(\sigma(x)) \cdot \textrm{arcsinh}(x)$ & 
    $72.51 {\scriptscriptstyle \pm 0.30}$ \\
    $-\textrm{Swish}(\textrm{Swish}(\alpha x))$ & 
    $\bm{72.49} {\scriptscriptstyle \pm 0.55}$ \\
    $-\textrm{Swish}(\textrm{Swish}(x))$ & 
    $71.97 {\scriptscriptstyle \pm 0.22}$ \\
    \midrule
    
    \multicolumn{2}{l}{\textbf{ResNet-v1-56}} \\
    $\alpha x - \beta \log(\sigma(\gamma x))$ & 
    $\bm{71.01} {\scriptscriptstyle \pm 0.64}$ \\
    $\alpha x - \log(\sigma(\beta x))$ & 
    $70.30 {\scriptscriptstyle \pm 0.58}$ \\
    $x - \log(\sigma(x))$ & 
    $69.29 {\scriptscriptstyle \pm 0.45}$ \\
    \midrule
    
    \multicolumn{2}{l}{\textbf{ResNet-v2-56}} \\
    $\min\{\log(\sigma(x)), \alpha \log(\sigma(\beta x))\}$ & 
    $75.19 {\scriptscriptstyle \pm 0.39}$ \\
    $\log(\sigma(x))$ & 
    $\bm{75.53} {\scriptscriptstyle \pm 0.37}$ \\
    \bottomrule
    \end{tabular}
    \end{adjustbox}
    \label{tab:disable_parameters}
\end{table}
\label{sec:effectofparameterization}

To understand the effect that parameterizing activation functions has on their performance, the specialized functions (Table \ref{tab:results}) were trained without them.  As Table \ref{tab:disable_parameters} shows, when parameters are removed, performance drops.  The function $\log(\sigma(x))$ is the only exception to this rule, but its high performance is not surprising, since it was previously discovered as a general activation function (Table \ref{tab:results}).  These results confirm that the learnable parameters contributed to the success of PANGAEA.

\subsection{Adding Parameters to Fixed Baseline Activation Functions}

\begin{table*}
    \centering
    \caption{CIFAR-100 test set accuracy aggregated over ten runs, shown as mean $\pm$ sample standard deviation.  Trivially parameterizing existing fixed activation functions does not substantially improve performance.  PANGAEA, on the other hand, utilizes evolutionary search to discover functional forms that are well suited to taking advantage of the parameters, leading to better performance.\newline
    }
    \centering
    \small
    \begin{adjustbox}{max width=\textwidth}
    \begin{tabular}{llll} \toprule
    & \textbf{WRN-10-4} & \textbf{ResNet-v1-56} & \textbf{ResNet-v2-56} \\ \midrule

    \textbf{Best Specialized Functions}\\
    $\log(\sigma(\alpha x)) \cdot \beta \textrm{arcsinh}(x)$ & 
    $\bm{73.20} {\scriptscriptstyle \pm 0.37}$ \\

    $\alpha x - \beta \log(\sigma(\gamma x))$ & &
    $\bm{71.01} {\scriptscriptstyle \pm 0.64}$ & \\
    
    $\textrm{Softplus}(\textrm{ELU}(x))$ & & &
    $\bm{75.61} {\scriptscriptstyle \pm 0.42}$ \\
    
    \midrule 
    
    \textbf{Parameterized Functions}\\
    $\alpha \textrm{ReLU}(\beta x)$ & 
    $71.96 {\scriptscriptstyle \pm 0.31}$ &
    $68.93 {\scriptscriptstyle \pm 0.22}$ &
    $73.52 {\scriptscriptstyle \pm 0.37}$ \\
    
    $\alpha \textrm{ELiSH}(\beta x)$ & 
    $01.00 {\scriptscriptstyle \pm 0.00}$&
    $01.00 {\scriptscriptstyle \pm 0.00}$&
    $73.94 {\scriptscriptstyle \pm 0.33}$\\
    
    $\alpha \textrm{ELU}(\beta x)$ & 
    $71.98 {\scriptscriptstyle \pm 0.24}$ &
    $69.06 {\scriptscriptstyle \pm 0.37}$ &
    $73.97 {\scriptscriptstyle \pm 0.45}$ \\
    
    $\alpha \textrm{GELU}(\beta x)$ & 
    $71.96 {\scriptscriptstyle \pm 0.34}$ &
    $69.39 {\scriptscriptstyle \pm 0.35}$ & 
    $73.83 {\scriptscriptstyle \pm 0.24}$ \\
    
    $\alpha \textrm{HardSigmoid}(\beta x)$ & 
    $66.70 {\scriptscriptstyle \pm 0.64}$ &
    $34.33 {\scriptscriptstyle \pm 6.53}$ &
    $65.10 {\scriptscriptstyle \pm 0.40}$ \\ 
    
    $\alpha \textrm{Leaky ReLU}(\beta x)$ & 
    $71.74 {\scriptscriptstyle \pm 0.39}$ &
    $69.11 {\scriptscriptstyle \pm 0.47}$ &
    $73.44 {\scriptscriptstyle \pm 0.29}$ \\ 
    
    $\alpha \textrm{Mish}(\beta x)$ & 
    $72.11 {\scriptscriptstyle \pm 0.31}$ &
    $69.51 {\scriptscriptstyle \pm 0.67}$ &
    $73.72 {\scriptscriptstyle \pm 0.32}$ \\
    
    $\alpha \textrm{SELU}(\beta x)$ & 
    $71.07 {\scriptscriptstyle \pm 0.33}$ &
    $68.05 {\scriptscriptstyle \pm 0.39}$ &
    $73.37 {\scriptscriptstyle \pm 0.38}$ \\
    
    $\alpha \textrm{sigmoid}(\beta x)$ & 
    $66.98 {\scriptscriptstyle \pm 0.66}$ &      
    $44.40 {\scriptscriptstyle \pm 2.62}$ &
    $66.98 {\scriptscriptstyle \pm 0.85}$ \\ 
    
    $\alpha \textrm{Softplus}(\beta x)$ & 
    $71.73 {\scriptscriptstyle \pm 0.31}$ &
    $68.84 {\scriptscriptstyle \pm 0.30}$ &
    $73.95 {\scriptscriptstyle \pm 0.37}$ \\ 
    
    $\alpha \textrm{Softsign}(\beta x)$ & 
    $62.12 {\scriptscriptstyle \pm 0.83}$ &
    $09.18  {\scriptscriptstyle \pm 13.75}$ &
    $68.87 {\scriptscriptstyle \pm 0.38}$ \\ 
    
    $\alpha \textrm{Swish}(\beta x)$ & 
    $72.26 {\scriptscriptstyle \pm 0.29}$ &
    $69.25 {\scriptscriptstyle \pm 0.28}$ &
    $73.93 {\scriptscriptstyle \pm 0.22}$ \\ 
    
    $\alpha \textrm{tanh}(\beta x)$ & 
    $63.55 {\scriptscriptstyle \pm 0.56}$ &
    $02.92 {\scriptscriptstyle \pm 6.07}$ &
    $69.55 {\scriptscriptstyle \pm 0.62}$ \\ 
    
    \bottomrule
    \end{tabular}
    \end{adjustbox}
    \label{tab:parametric_baseline}
\end{table*}
\label{sec:parametric_baseline}

As demonstrated in Tables \ref{tab:results} and \ref{tab:disable_parameters}, learnable parameters are an important component of PANGAEA.  An interesting question is whether accuracy can be increased simply by augmenting existing activation functions with learnable parameters.  Table \ref{tab:parametric_baseline} shows that this is not the case: trivially adding parameters to fixed activation functions does not reliably improve performance.  This experiment implies that certain functional forms are better suited to taking advantage of parameterization than others.  By utilizing evolutionary search, PANGAEA is able to discover these functional forms automatically.

\subsection{Scaling Up to Larger Networks} 
\label{sec:scalingup}

PANGAEA discovered specialized activation functions for WRN-10-4, ResNet-v1-56, and ResNet-v2-56.  Table \ref{tab:scale_up} shows the performance of these activation functions when paired with the larger WRN-16-8, ResNet-v1-110, and ResNet-v2-110 architectures.  Due to time constraints, ReLU is the only baseline activation function in these experiments. 

Two of the three functions discovered for WRN-10-4 outperform ReLU with WRN-16-8, and all three functions discovered for ResNet-v2-56 outperform ReLU with ResNet-v2-110.  Interestingly, ReLU achieves the highest accuracy for ResNet-v1-110, where activation functions are part of the skip connections, but not for ResNet-v2-110, where they are not. Thus, it is easier to achieve high performance with specialized activation functions on very deep architectures when they are not confounded by skip connections.  Notably, ResNet-v2-110 with $\textrm{Softplus}(\textrm{ELU}(x))$ performs comparably to the much larger ResNet-v2-1001 with ReLU (77.14 vs.\ 77.29, as reported by \citet{he2016identity}). 

Evolving novel activation functions can be computationally expensive.  The results in Table \ref{tab:scale_up} suggest that it is possible to reduce this cost by evolving activation functions for smaller architectures, and then using the discovered functions with larger architectures.

\begin{table}
    \centering
    \caption{Specialized activation functions discovered for WRN-10-4, ResNet-v1-56, and ResNet-v2-56 are evaluated on larger versions of those architectures: WRN-16-8, ResNet-v1-110, and ResNet-v2-110, respectively.  CIFAR-100 test accuracy is reported as mean $\pm$ sample standard deviation across three runs.  Specialized activation functions successfully transfer to WRN-16-8 and ResNet-v2-110, outperforming ReLU.
    \newline}
    \small
    \begin{adjustbox}{max width=\linewidth}
    \begin{tabular}{ll} 
        \toprule
         \textbf{WRN-16-8}  \\ 
         $\log(\sigma(\alpha x)) \cdot \beta \textrm{arcsinh}(x)$ & $\bm{78.36} {\scriptscriptstyle \pm 0.17}$ \\
         $\log(\sigma(\alpha x)) \cdot \textrm{arcsinh}(x)$ & $78.34 {\scriptscriptstyle \pm 0.20}$ \\
         $-\textrm{Swish}(\textrm{Swish}(\alpha x))$ & $78.00 {\scriptscriptstyle \pm 0.35}$ \\ 
         ReLU & $78.15 {\scriptscriptstyle \pm 0.03}$\\
        \midrule
         \textbf{ResNet-v1-110}  \\ 
         $\alpha x - \beta \log(\sigma(\gamma x))$ & $70.85 {\scriptscriptstyle \pm 0.50}$ \\
         $\alpha x - \log(\sigma(\beta x))$ & $70.34 {\scriptscriptstyle \pm 0.60}$ \\
         $\max\{\textrm{Swish}(x), 0\}$ & $70.36 {\scriptscriptstyle \pm 0.56}$ \\ 
         ReLU & $\bm{71.23} {\scriptscriptstyle \pm 0.25}$ \\
        \midrule 
         \textbf{ResNet-v2-110}  \\ 
         $\textrm{Softplus}(\textrm{ELU}(x))$ & $\bm{77.14} {\scriptscriptstyle \pm 0.38}$ \\
         $\min\{\log(\sigma(x)), \alpha \log(\sigma(\beta x))\}$ & $76.93 {\scriptscriptstyle \pm 0.19}$ \\
         $\textrm{SELU}(\textrm{Swish}(x))$ & $76.96 {\scriptscriptstyle \pm 0.14}$ \\ 
         ReLU & $76.34 {\scriptscriptstyle \pm 0.11}$ \\
        \bottomrule
    \end{tabular}
    \end{adjustbox}
    \label{tab:scale_up}
\end{table}

\subsection{A New Task: All-CNN-C on CIFAR-10} 
\begin{table}
    \centering
    \caption{All-CNN-C accuracy with different activation functions on CIFAR-10, shown as mean $\pm$ sample standard deviation across ten runs.  PANGAEA improves performance significantly also with this different architecture and task.\newline}
    \small
    \begin{adjustbox}{max width=\linewidth}
    \begin{tabular}{ll}
    \toprule
    $\alpha\textrm{ReLU}(\beta |\textrm{ReLU}(\gamma x)|)$  & $\bm{92.77} {\scriptscriptstyle \pm 0.13}$\\
    $\alpha \textrm{Swish}(x) \cdot \cosh(\beta)$ & $92.66 {\scriptscriptstyle \pm 0.08}$\\
    $\alpha \textrm{Swish}(\beta x)$ & $76.15 {\scriptscriptstyle \pm 34.86}$\\
    ReLU & $88.47 {\scriptscriptstyle \pm 0.14}$\\
    \bottomrule
    \end{tabular}
    \end{adjustbox}
    \label{tab:allcnnc}
\end{table}
\label{sec:allcnnc}
To verify that PANGAEA is effective with different datasets and types of architectures, activation functions were evolved for the All-CNN-C \citep{springenberg2015striving} architecture on the CIFAR-10 dataset.  All-CNN-C is quite distinct from the architectures considered above: It contains only convolutional layers, activation functions, and a global average pooling layer, but it does not have residual connections.

As shown in Table \ref{tab:allcnnc}, PANGAEA improves significantly over ReLU in this setting as well.  The accuracy improvement from 88.47\% to 92.77\% corresponds to an impressive 37.29\% reduction in the error rate.  This experiment provides further evidence that PANGAEA can improve performance for different architectures and tasks.

\subsection{Training Dynamics of Evolved Activation Functions}
\label{sec:dynamics}

PANGAEA discovers activation functions that improve accuracy over baseline functions.  An interesting question is: What mechanisms do these evolved functions use in order to achieve better performance?  By examining training and validation curves qualitatively for different activation functions, it appears that some functions ease optimization, while others improve performance through implicit regularization.

\begin{figure}
    \centering
    \begin{tikzpicture}
        \draw (0, 0) node[anchor=north west, inner sep=0, align=left] {
            \includegraphics[trim={0 0 0 11em}, clip, width=0.48\textwidth]{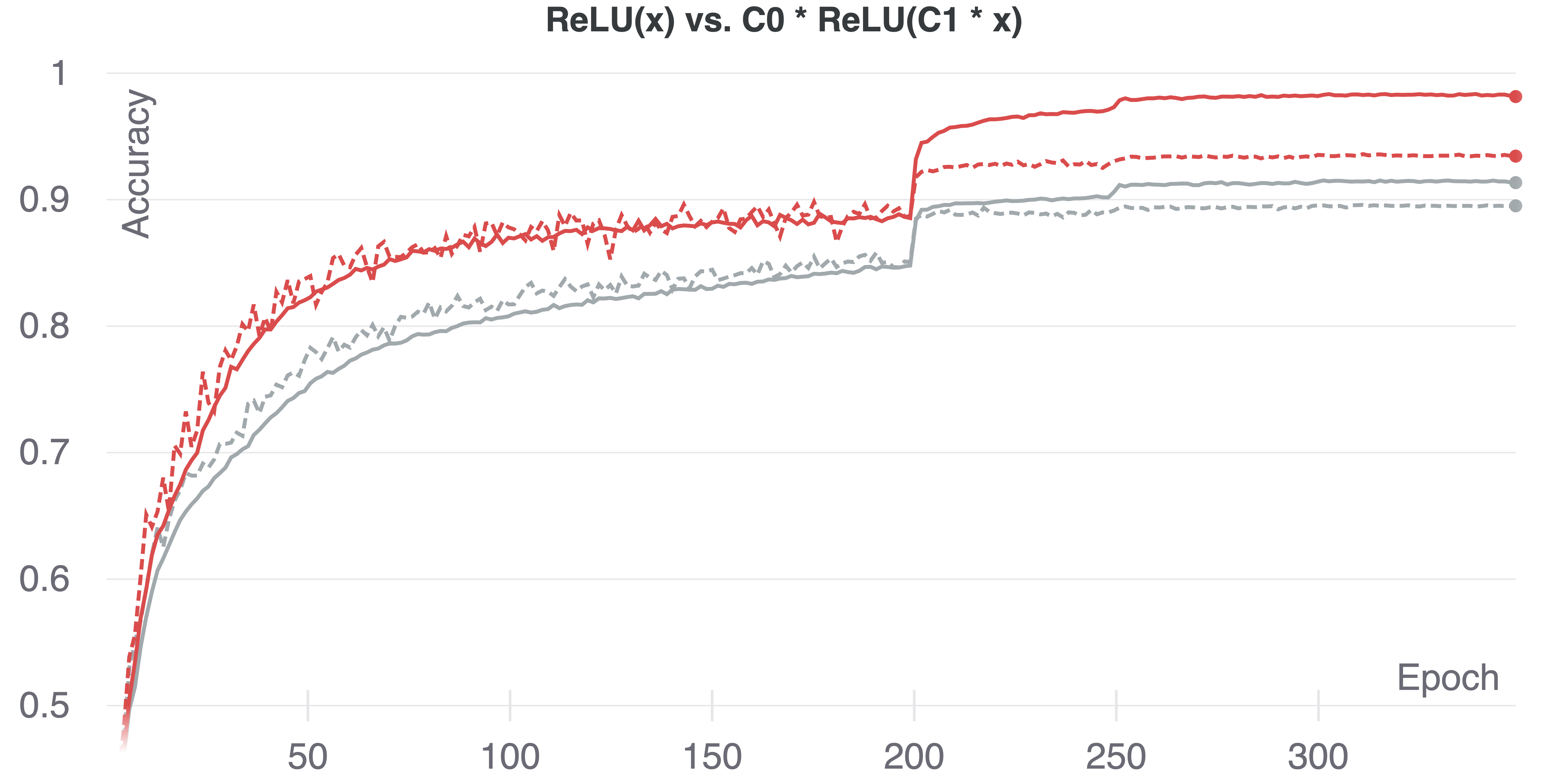} };
        \draw (0, 0.5) node[anchor=north west, inner sep=0, align=left] {\small \textbf{(a):} $\textrm{ReLU}(x) \textrm{ vs. } \alpha \cdot \textrm{ReLU}(\beta \cdot x)$ };
    \end{tikzpicture}
    \begin{tikzpicture}
        \draw (0, 0) node[anchor=north west, inner sep=0, align=left] {
            \includegraphics[trim={0 0 0 11em}, clip, width=0.48\textwidth]{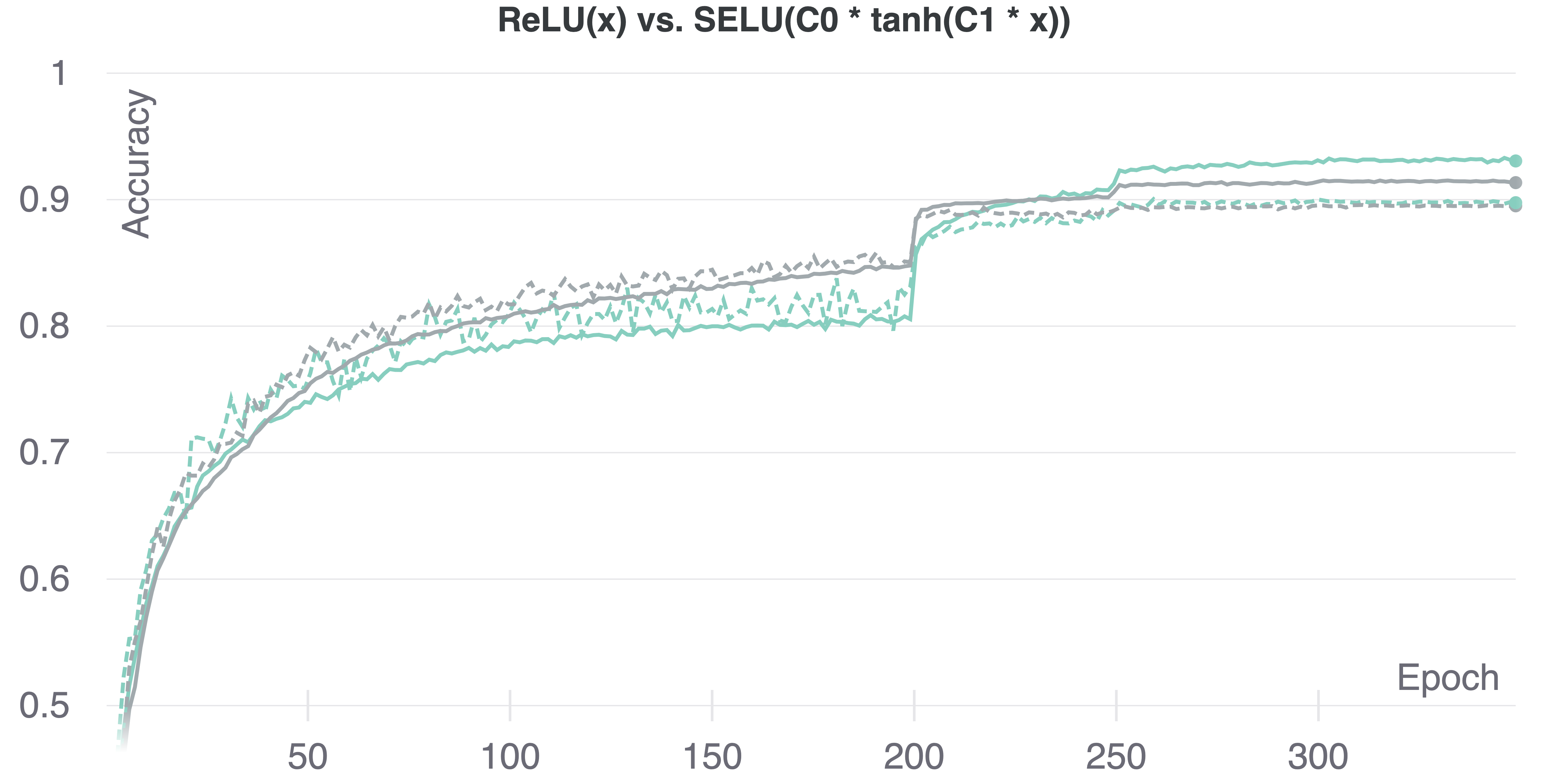} };
        \draw (0, 0.5) node[anchor=north west, inner sep=0, align=left] {\small \textbf{(b):} $\textrm{ReLU}(x) \textrm{ vs. } \textrm{SELU}(\alpha \cdot \tanh(\beta \cdot x))$ };
    \end{tikzpicture}\\[1em]
    \begin{tikzpicture}
        \draw (0, 0) node[anchor=north west, inner sep=0, align=left] {
            \includegraphics[trim={0 0 0 11em}, clip, width=0.48\textwidth]{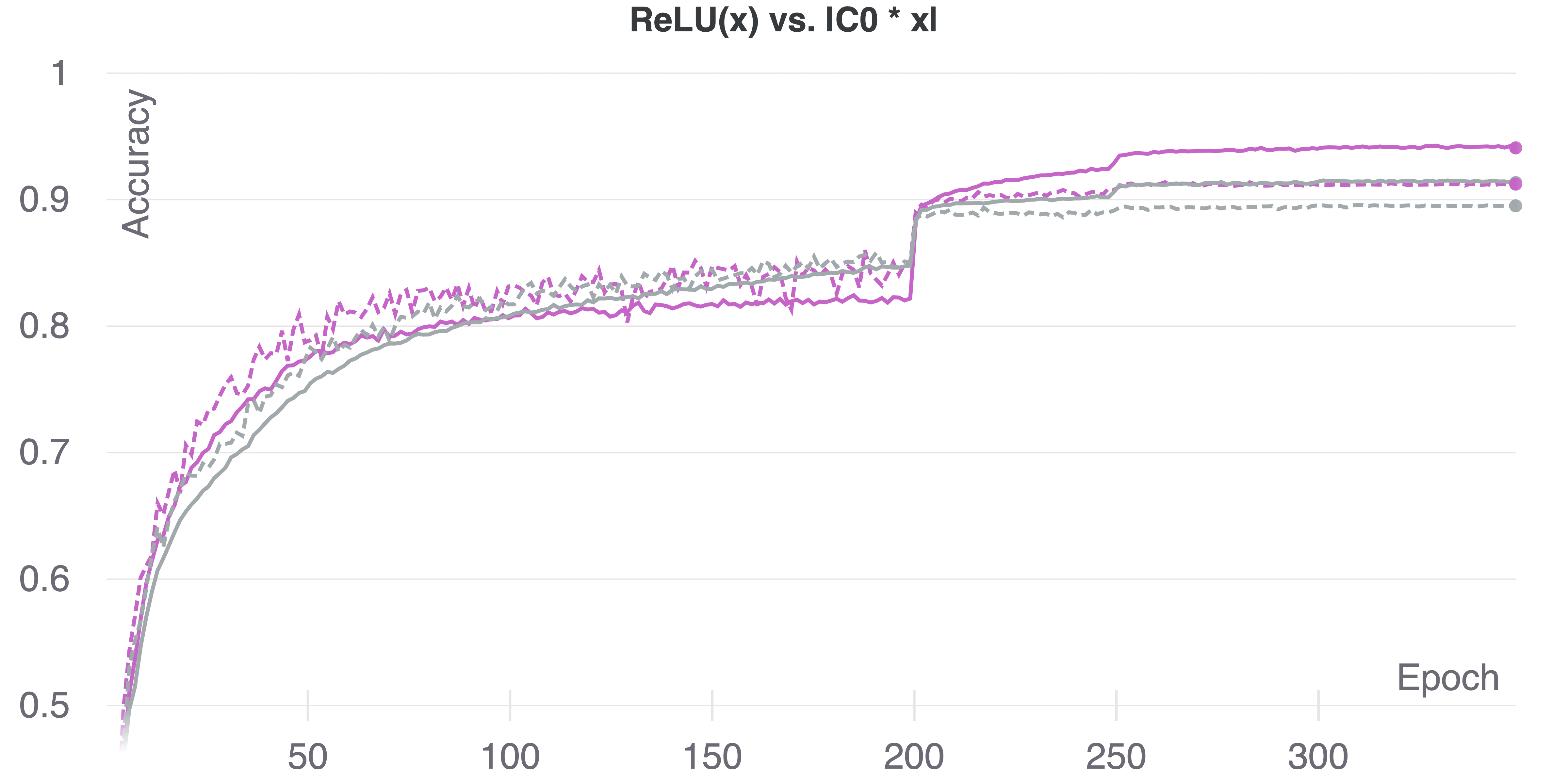} };
        \draw (0, 0.5) node[anchor=north west, inner sep=0, align=left] {\small \textbf{(c):} $\textrm{ReLU}(x) \textrm{ vs. } |\alpha \cdot x|$};
    \end{tikzpicture}
    \begin{tikzpicture}
        \draw (0, 0) node[anchor=north west, inner sep=0, align=left] {
            \includegraphics[trim={0 0 0 11em}, clip, width=0.48\textwidth]{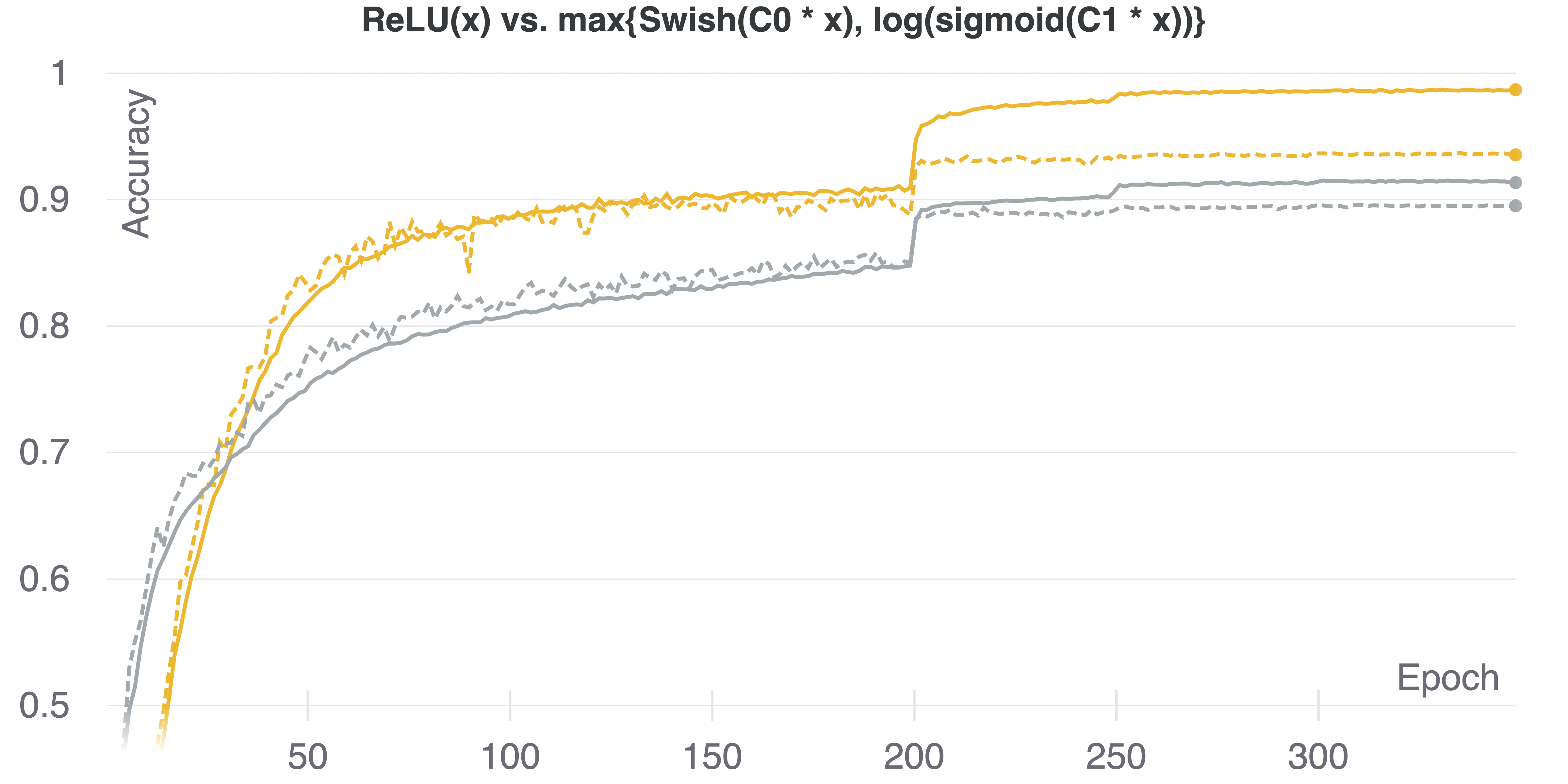} };
        \draw (0, 0.5) node[anchor=north west, inner sep=0, align=left] {\small \textbf{(d):} $\textrm{ReLU}(x) \textrm{ vs. } \max\{ \textrm{Swish}(\alpha \cdot x), \log(\sigma(\beta \cdot x))\}$};
    \end{tikzpicture}
    \caption{Training curves of All-CNN-C on CIFAR-10 with different activation functions.  Accuracy with ReLU is shown in gray, and accuracy with the discovered functions in different colors.  The training accuracy is shown as a solid line, and validation accuracy as a dashed line.  All of these discovered functions outperformed ReLU, but the training curves show different behavior in each case. In (a) both curves are above those of ReLU the whole time, suggesting ease of learning. In (b), they exceed those of ReLU only in the end, suggesting early regularization. More complex profiles (such as those in (c) and (d)) are also observed, suggesting that their combinations are possible as well.}
    \label{fig:allcnnc_training_curves}
\end{figure}
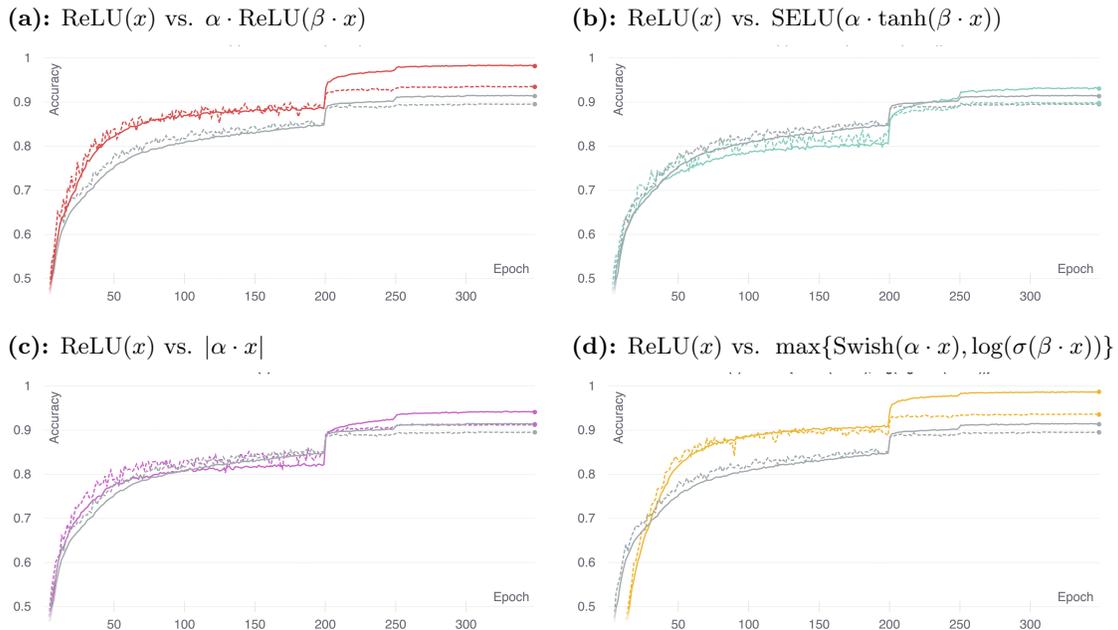

For instance, Figure \ref{fig:allcnnc_training_curves} shows training and validation curves for the All-CNN-C architecture and four discovered activation functions, plus ReLU. With All-CNN-C, the learning rate starts at 0.01 and decreases by a factor of 0.1 after epochs 200, 250, and 300, with training ending at epoch 350.  With some discovered activation functions, the training and validation curves are consistently higher than the curves from ReLU across all epochs of training, indicating easier optimization (Figure \ref{fig:allcnnc_training_curves}a).  With other activation functions, the training and validation curves actually remain lower than those from ReLU until the final stage in the learning rate schedule, suggesting implicit regularization (Figure \ref{fig:allcnnc_training_curves}b). In such cases, the network is learning difficult patterns in the early stages of training; in contrast, the ReLU model memorizes simpler patterns, which leads to early gains but difficulty generalizing later on \citep{li2019towards}.

Even more complex behavior can be observed in some cases.  For example, some discovered functions have training and validation curves that start out higher than those from ReLU, plateau to a lower value, but then again surpass those from ReLU at later stages in the learning rate schedule (Figure \ref{fig:allcnnc_training_curves}c).  Others have curves that start out lower than those from ReLU, but surpass it within a few dozen epochs (Figure \ref{fig:allcnnc_training_curves}d). Such diverse behavior suggests that these mechanisms can be combined in complex ways. Thus, the plots in Figure \ref{fig:allcnnc_training_curves} suggest that in the future it may be feasible to search for activation functions with specific properties depending on the task at hand.  For example, a larger network may benefit from an activation function that implicitly regularizes it, while a smaller network may be better suited to an activation function that eases optimization. 

\section{Evaluating Robustness: Experiments on Reliability, Flexibility, and Efficiency}

This section demonstrates robustness of PANGAEA with three experiments.  In the first experiment, two independent PANGAEA processes were run from scratch.  The processes discovered similarly powerful activation functions, demonstrating that PANGAEA reliably discovers good activation functions each time it is run.  In the second experiment, variations of PANGAEA with per-layer and per-neuron learnable parameters were run, to complement the original PANGAEA with per-channel parameters.  The results show that PANGAEA is effective with all three variations.  Third, statistics from activation functions across the per-layer, two per-channel, and the per-neuron variations were aggregated to demonstrate computational efficiency of PANGAEA.  Every PANGAEA run discovers an activation function that outperforms ReLU early on in the search process, before much compute is spent; they also eventually discover activation functions that perform much better and train almost as quickly as ReLU.

\subsection{Reliability of PANGAEA}

PANGAEA is inherently a stochastic process.  Therefore, an important question is whether PANGAEA can discover good activation functions reliably every time it is run.  To answer this question, PANGAEA was run from scratch on ResNet-v1-56 independently two times.  These runs utilized per-channel parameters, and were identical to the original PANGAEA run, except they were allowed to evaluate up to $C=2{,}000$ activation functions instead of $C=1{,}000$ from the original run.  They were run on ResNet-v1-56 since it proved to be the most difficult architecture to optimize (Table \ref{tab:results}).  

\begin{table}
    \centering
    \begin{tabular}{ll}
        \toprule
        \textbf{Per-layer PANGAEA}\\
        $\max\{\textrm{Swish}(x) , x\}$ & $71.00 {\scriptscriptstyle \pm 0.28}$\\
        $\max\{x, \alpha \cdot \log(\sigma(\textrm{SELU}(x)))\}$ & $70.76 {\scriptscriptstyle \pm 0.29}$\\
        $\max\{\alpha \cdot \max\{\beta \cdot \textrm{ReLU}(\textrm{arcsinh}(x)), x\}, \max\{\gamma \cdot \textrm{ReLU}(\textrm{arcsinh}(x)), x\}\}$ & $70.63 {\scriptscriptstyle \pm 0.35}$\\
        \midrule
        
        \textbf{Original Per-channel PANGAEA Run}\\
        $\alpha x - \beta \log(\sigma(\gamma x))$ & 
        $71.01 {\scriptscriptstyle \pm 0.64}$ \\
        $\alpha x - \log(\sigma(\beta x))$ & 
        $70.30 {\scriptscriptstyle \pm 0.58}$ \\
        $\max\{\textrm{Swish}(x), 0\}$ & 
        $69.43 {\scriptscriptstyle \pm 0.69}$ \\
        \midrule
        
        \textbf{Additional Per-channel PANGAEA Run 1} \\
        $\max \{ \min\{ \alpha \cdot x , \textrm{ELU}(x)\} , 0 \}$ & $70.53 {\scriptscriptstyle \pm 0.31}$\\ 
        $\alpha \cdot \max\left\{\beta \cdot \textrm{ReLU}\left(\frac{\textrm{Swish}(x)}{\gamma}\right), x\right\}$ & $70.52 {\scriptscriptstyle \pm 0.39}$\\
        $\max\{ \textrm{ReLU}(\textrm{Swish}(\alpha \cdot x)), \beta \cdot x\}$ & $70.44 {\scriptscriptstyle \pm 0.44}$\\
        \midrule
        
        \textbf{Additional Per-channel PANGAEA Run 2} \\
        $\max\{ \textrm{Swish}(\alpha \cdot x), x \}$ & $71.03 {\scriptscriptstyle \pm 0.40}$\\
        $\max\{ \textrm{Swish}(x), \textrm{arcsinh}(\alpha \cdot \beta \cdot x)\}$ & $70.52 {\scriptscriptstyle \pm 0.35}$\\ 
        $\max\{ \textrm{Swish}(x), \textrm{arcsinh}(\alpha \cdot x)\}$ & $70.41 {\scriptscriptstyle \pm 0.38}$\\
        \midrule
        
        \textbf{Per-neuron PANGAEA}\\
        $\alpha \cdot \max\{\beta \cdot \textrm{Swish}(\gamma \cdot x), \textrm{Swish}(x)\}$ & $71.25 {\scriptscriptstyle \pm 0.35}$\\
        $\alpha \cdot x - (\beta \cdot \textrm{Swish}(\gamma \cdot x))$ & $71.23 {\scriptscriptstyle \pm 0.18}$\\
        $\textrm{ELU}(\textrm{ELU}(\alpha \cdot x) + \log(\sigma(0)))$ & $71.20 {\scriptscriptstyle \pm 0.25}$\\
        \midrule
        
        \textbf{Random Sample $(n=500)$}\\
        Per-layer $\max\{\alpha \cdot \textrm{Swish}(\beta \cdot x), \textrm{Softplus}(x)\}$ & $70.32$ \\
        Per-channel $\alpha \cdot x^2$ & $70.91$\\
        Per-neuron $\textrm{bessel\_i0e}(|x|) + \alpha \cdot |x|$ & $\bm{71.66}$ \\
        \midrule
        
        ReLU & $69.64 {\scriptscriptstyle \pm 0.65}$\\
        \bottomrule
    \end{tabular}
    \caption{Performance of the best activation functions from multiple PANGAEA runs with ResNet-v1-56.  CIFAR-100 test accuracy is shown as the mean $\pm$ sample standard deviation across three runs.  The three independent per-channel runs produce activation functions of similar performance, demonstrating the reliability of PANGAEA.  PANGAEA also discovers good activation functions with per-layer or per-neuron parameters, showing its flexibility. The very best per-layer and per-neuron functions are difficult to find, suggesting that their distribution is long-tailed.}
    \label{tab:resnetv1_lcn}
\end{table}

There are multiple ways to analyze the similarity of the two PANGAEA runs.  First, a simple and relevant metric is to look at the test accuracies of the functions discovered by each search.  Table~\ref{tab:resnetv1_lcn} shows that both PANGAEA runs discovered multiple good activation functions.  The accuracies achieved by these functions are substantially higher than those achieved by ReLU. Most importantly, although the functions themselves are different, they resulted in similar accuracies as functions from the original PANGAEA run.  

\begin{figure}
    \centering
    \includegraphics{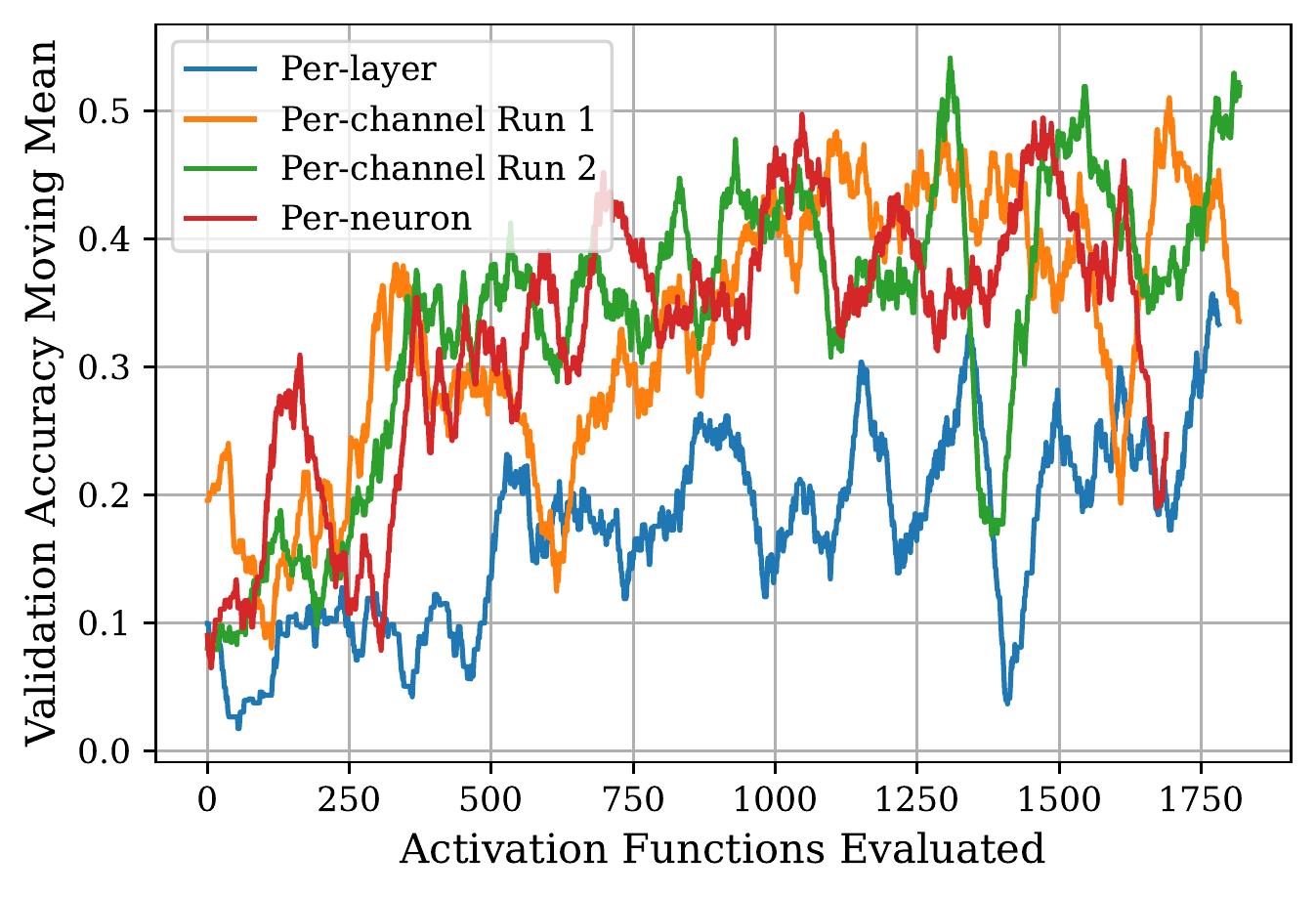}
    \caption{Average population fitness across time for four independent PANGAEA runs.  The plots show the average validation accuracy achieved with the 64 most recently evaluated activation functions at any given time in the search process.  All four PANGAEA runs gradually discover better activation functions as they explore the search space, with the per-layer run slightly below the others.  Importantly, the two per-channel PANGAEA runs progress at similar rates, demonstrating the reliability of PANGAEA.}
    \label{fig:moving_acc_lcn}
\end{figure}

A second way is to compare the time course of discovery. To this end, Figure \ref{fig:moving_acc_lcn} shows how the populations of $P=64$ activation functions improved over time in the two PANGAEA runs.  In both cases, the initial functions are relatively poor.  As evolution progresses, both runs discover better functions at similar rates.  This result shows that in addition to the final results, the PANGAEA process as a whole is stable and reliable.

\begin{figure}
    \centering
    \includegraphics[width=\linewidth]{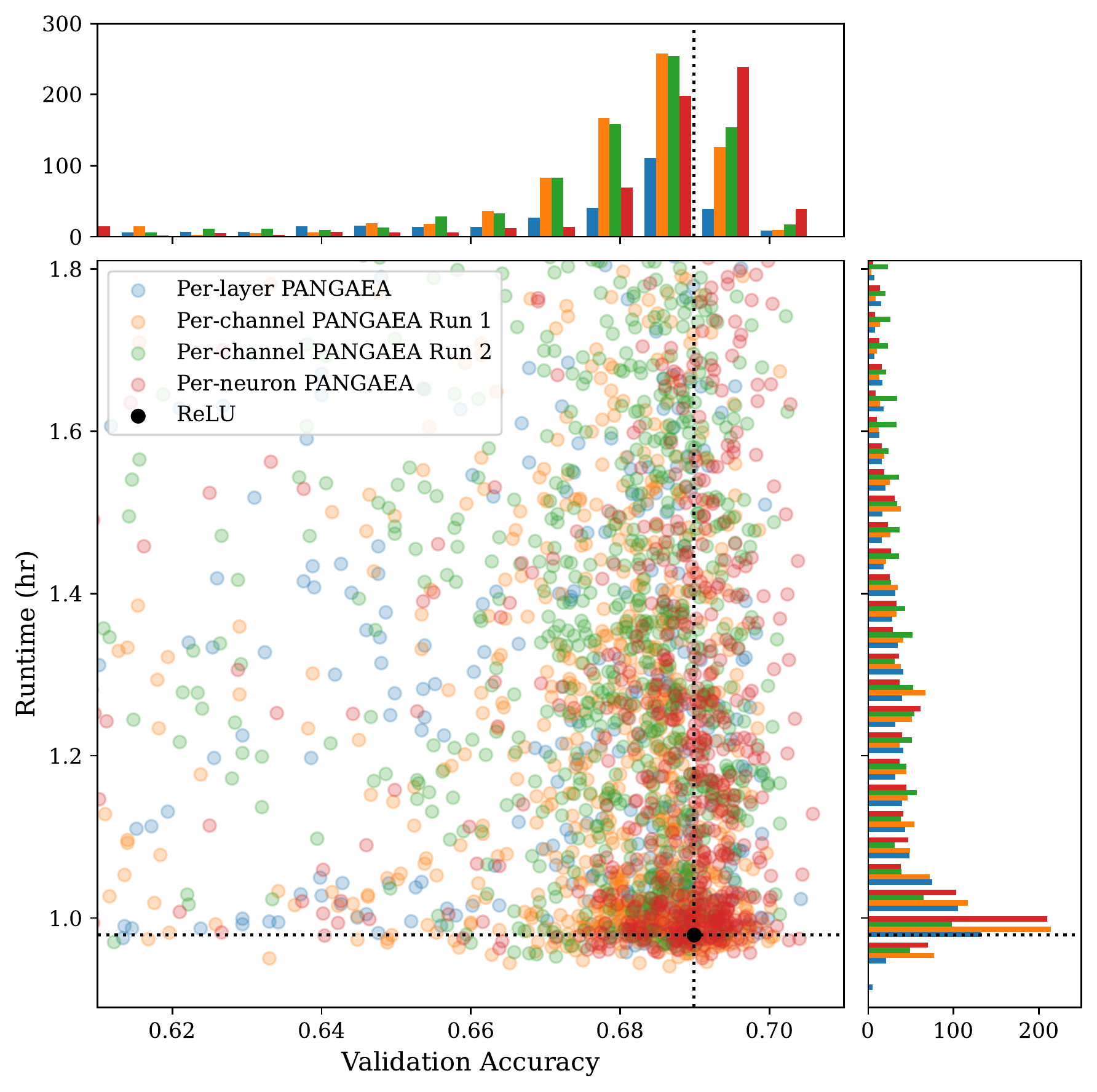}
    \caption{Fitness (validation accuracy) and compute cost (runtime in hours) among all activation functions considered in four independent PANGAEA processes on ResNet-v1-56.  Each point represents a different activation function.  The distribution of fitness and compute cost are shown in histograms on the top and right, respectively.  All PANGAEA variants explore activation functions of similar complexity and reliably discover many novel functions that outperform ReLU.}
    \label{fig:scatter_hist_lcn}
\end{figure}

A third way is to compare the complexity, i.e.\ time it takes to train the network with the discovered functions. Figure~\ref{fig:scatter_hist_lcn} shows the cost of all of the activation functions considered throughout each PANGAEA run. The runtimes of activation functions are comparable across different validation accuracies, suggesting that both runs discovered functions of similar complexity.

In sum, the two PANGAEA runs produced comparable final results, progressed at comparable rates, and searched through functions of similar complexity.  These results suggest that PANGAEA is a reliable process that can consistently outperform baseline activation functions.

\subsection{Parameters: per-layer, per-channel, or per-neuron?}

Learnable parameters in activation functions can be per-layer, per-channel, or per-neuron. It is not clear which setting is the best.  For example, per-neuron parameters are the default setting in the TensorFlow implementation of PReLU \cite{he2015delving}.  However, \citet{he2015delving} also experimented with per-channel and per-layer implementations.  Further preliminary experiments for this paper (Table~\ref{tab:prelu_lcn}) suggest that per-neuron PReLU is best for WRN-10-4 and ResNet-v2-56, but this setting is the worst for ResNet-v1-56, which benefits most from per-layer parameters.  

Similarly, no clear trends were observed in preliminary PANGAEA experiments.  For some activation functions and architectures per-neuron parameters were beneficial, presumably due to the added expressivity of each neuron learning its own optimal activation function.  In other cases per-layer was better, possibly due to an implicit regularization effect caused by all neurons within a layer using the same activation function.  As a compromise between the expressivity and regularization of these two strategies, per-channel parameters were utilized in the main experiments.  However, the preliminary results suggest that performance may be further optimized by specializing the parameter setting to the activation function and to the architecture.

\begin{table}[]
    \centering
    \begin{tabular}{llll}
        \toprule
        & \textbf{WRN-10-4} & \textbf{ResNet-v1-56} & \textbf{ResNet-v2-56} \\
        \midrule
        Per-layer PReLU & $71.92 {\scriptscriptstyle \pm 0.41}$ & $\bm{71.40} {\scriptscriptstyle \pm 0.59}$ & $73.54 {\scriptscriptstyle \pm 0.21}$ \\
        Per-channel PReLU & $71.15 {\scriptscriptstyle \pm 0.41}$ & $71.25 {\scriptscriptstyle \pm 0.54}$ & $74.52 {\scriptscriptstyle \pm 0.24}$ \\
        Per-neuron PReLU & $\bm{72.23} {\scriptscriptstyle \pm 0.37}$ & $69.77 {\scriptscriptstyle \pm 0.40}$ & $\bm{75.10} {\scriptscriptstyle \pm 0.53}$ \\
        \bottomrule
    \end{tabular}
    \caption{CIFAR-100 test accuracy with different architectures and PReLU variants reported as mean $\pm$ sample standard deviation across ten runs.  Per-neuron PReLU gets the best performance on WRN-10-4 and ResNet-v2-56, but per-layer PReLU is the best for ResNet-v1-56.  }
    \label{tab:prelu_lcn}
\end{table}

To explore this idea further, per-layer and per-neuron versions of PANGAEA were run from scratch on ResNet-v1-56. Both of these PANGAEA runs produced good activation functions that outperformed ReLU substantially (Table \ref{tab:resnetv1_lcn}). Interestingly, although per-layer PReLU outperformed per-neuron PReLU with ResNet-v1-56 (Table~\ref{tab:prelu_lcn}), PANGAEA performed the best in the per-neuron setting (Table \ref{tab:resnetv1_lcn}).  Indeed, although the per-layer PANGAEA runs still discovered good activation functions, their average performance during search was often lower than that of the per-channel or per-neuron variants (Figure \ref{fig:moving_acc_lcn}).  These findings suggest that the distribution of per-layer activation functions may be long-tailed: Powerful per-layer activation functions do exist, but they may be more difficult to discover compared to per-channel or per-neuron activation functions.

In order to separate the search space from the search algorithm, in a further experiment 500 per-layer, per-channel, and per-neuron activation functions were randomly created and trained once with ResNet-v1-56 (the functions were first initialized randomly as shown in Figure \ref{fig:initialization}, and then mutated randomly three times as shown in Figure \ref{fig:mutation}).  The best activation functions from these random samples are included in Table \ref{tab:resnetv1_lcn}.  The best per-layer activation function outperformed ReLU by a large margin, but was not as powerful as those discovered with PANGAEA; the best per-neuron function outperformed all other variants.  These results thus suggest that the distribution of good per-neuron functions may be long-tailed as well, but at a higher level of performance than per-layer and per-channel functions.

In sum, although more research is needed to discover a principled way to select per-layer, per-channel, or per-neuron parameters in a given situation, PANGAEA is flexible enough to discover good functions for all three of these cases.

\subsection{Efficiency of PANGAEA}

PANGAEA's computational efficiency needs to be evaluated from two perspectives.  First, how much compute is necessary to find good activation functions?  Second, once a good activation function is found, how much more expensive is it to use it in a network compared to a baseline function like ReLU?  This section aggregates data from the per-layer, the two per-channel, and the per-neuron PANGAEA runs to demonstrate that PANGAEA is surprisingly efficient in both respects.

\begin{figure}
    \centering
    \includegraphics{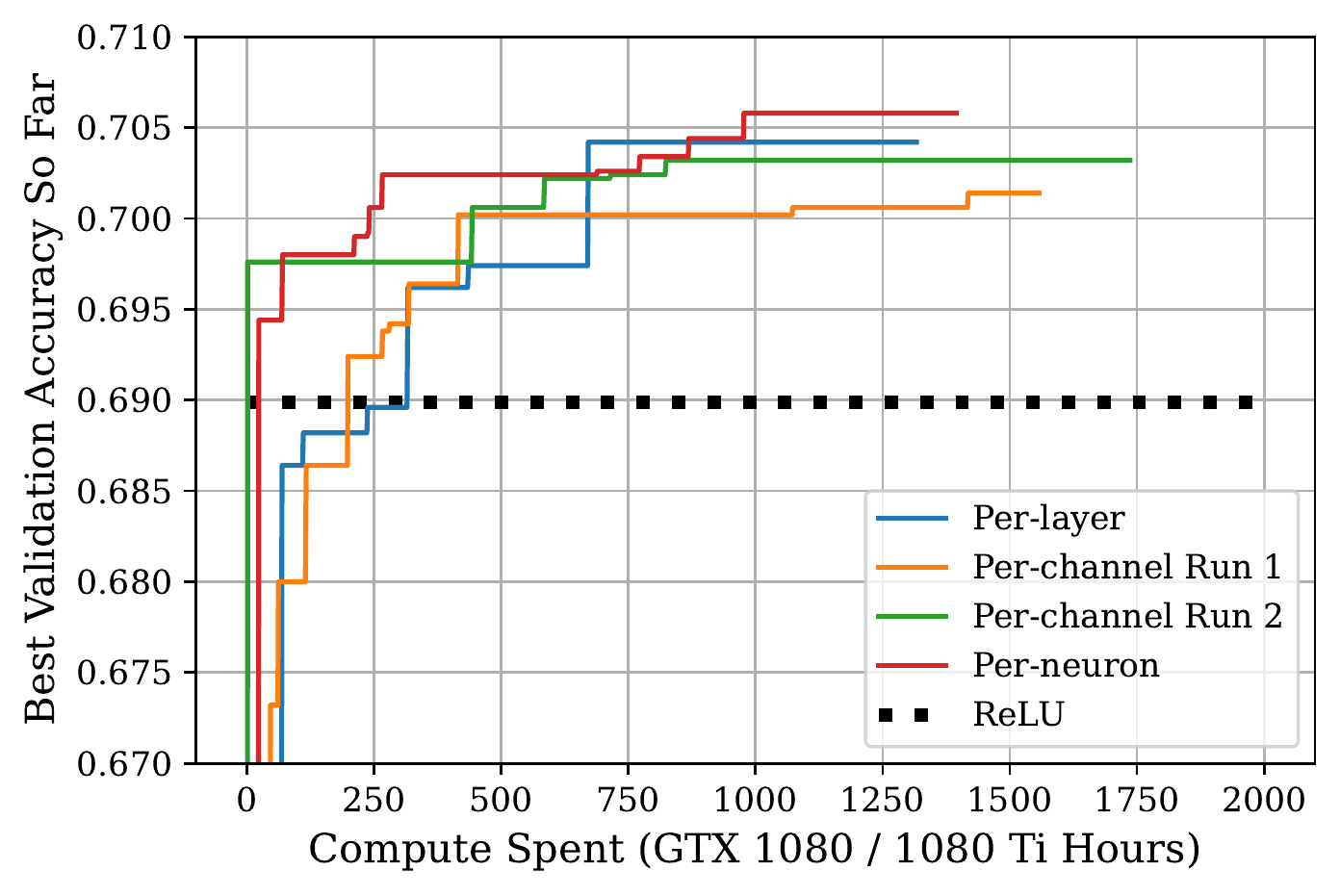}
    \caption{Computational efficiency of PANGAEA.  The plot shows the performance of the best activation function discovered so far ($y$-axis) after a given amount of compute was spent ($x$-axis).  All four PANGAEA runs discover activation functions that outperform ReLU with relatively little compute, demonstrating the efficiency of PANGAEA.  If even better performance is needed, additional compute can be spent.}
    \label{fig:compute_spent}
\end{figure}

First, Figure \ref{fig:compute_spent} shows how the four PANGAEA runs discovered better activation functions with increasingly more compute.  All four runs discovered an activation function that outperformed ReLU relatively early in the search.  Because some activation functions are unstable and cause training to fail, they require negligible compute.  Computational resources can instead be focused on functions that appear promising.  The implications of Figure \ref{fig:compute_spent} are that in practice, PANGAEA can be used to improve over a baseline activation function relatively cheaply.  If better performance is needed, additional compute can be used to continue the search until the desired performance is achieved.

Second, Figure \ref{fig:scatter_hist_lcn} shows the distribution of accuracy and compute cost of all activation functions evaluated in all four PANGAEA runs on ResNet-v1-56.  Each point in the scatter plot represents a unique activation function discovered in one of the searches, and the distribution of accuracies and compute costs are shown as histograms above and to the side of the main plot.  The results confirm earlier conclusions: All strategies find many activation functions that beat ReLU, and per-neuron PANGAEA discovers more high-performing functions than per-channel or per-layer PANGAEA. The total amount of time it took to train the architecture with the given activation function is shown as ``runtime'' in the vertical axis.  Interestingly, there is a wide range in this metric: some activation functions are significantly more expensive than ReLU, while others are essentially the same. Thus, the figure shows that there exist plenty of activation functions that significantly beat ReLU but do not incur a significant computational overhead.  This distribution also suggests that a multi-objective approach that optimizes for both accuracy and computational cost simultaneously could be effective.



\section{Future Work}

It is difficult	to select an appropriate activation function for a
given architecture because the activation function, network 
topology, and training setup interact in
complex ways.  It is especially promising that PANGAEA discovered activation functions that significantly outperformed
the baselines, since the architectures and training setups were standard and developed with ReLU. A compelling research direction is to jointly optimize the architecture, training setup, and activation function.

More specifically, there has been significant recent research in
automatically discovering the architecture of neural networks through
gradient-based, reinforcement learning, or neuroevolutionary methods
\citep{elsken2019neural, wistuba2019survey, real2019regularized}.  In
related work, evolution was used discover novel loss functions
automatically
\citep{gonzalez2019improved,gonzalez2020evolving,liang2020population},
outperforming the standard cross entropy loss.	In the future, it may
be possible to optimize many of these aspects of neural network design
jointly.  Just as new activation functions improve the accuracy of
existing network architectures, it is likely that different
architectures will be discovered when the activation function is not
ReLU.  One such example is EfficientNet \citep{tan2019efficientnet},
which achieved state-of-the-art accuracy for ImageNet \citep{deng2009imagenet} using the Swish activation function
\citep{DBLP:conf/iclr/RamachandranZL18, elfwing2018sigmoid}.
Coevolution of activation functions, topologies, loss
functions, and possibly other aspects of neural network design
could allow taking advantage of interactions between them, leading to
further improvements in	the future.  

Similarly, there can be multiple properties of an activation function
that make it useful in different scenarios.  PANGAEA optimized for
activation functions that lead to high accuracy.  Another promising area
of future work is to search for activation functions that are also efficient to compute (Figure \ref{fig:scatter_hist_lcn}), improve adversarial robustness \cite{xie2020smooth}, stabilize training \cite{selu}, or meet other objectives like easing optimization or providing implicit regularization \cite{li2019towards}. These functions could be discovered with multi-objective optimization, or through other methods like a carefully designed search space or adding regularization to the parameters of the activation functions.

\section{Conclusion}

This paper introduced PANGAEA, a technique for automatically designing
novel, high-performing, parametric activation functions.  PANGAEA
builds a synergy of two different optimization processes: evolutionary
population-based search	for the	general	form, and
gradient descent-based fine-tuning of the parameters of	the activation
function.  Compared to previous studies, the search space is extended
to include deeper and more complex functional forms, including ones
unlikely to be discovered by humans.  The parameters are adapted during 
training and are different in different locations of the
architecture, thus customizing the functions over both time and space.
PANGAEA is able to discover general activation functions that perform
well across architectures, and specialized functions taking advantage of
a particular architecture, significantly outperforming previously proposed
activation functions in both cases.
It is thus a promising step towards automatic configuration of neural networks.

\section*{Acknowledgements}

The authors acknowledge the Texas Advanced Computing Center (TACC) at The University of Texas at Austin for providing HPC resources that have contributed to the research results reported within this paper.  
This research did not receive any specific grant from funding agencies in the public, commercial, or not-for-profit sectors.

\bibliography{references}
\bibliographystyle{abbrvnat}

\appendix

\section{Training Details}
\label{ap:details}

\subsection{Wide Residual Network (WRN-10-4)}

When measuring final performance after evolution, the standard WRN setup is used; all ReLU activations in WRN-10-4 are replaced with the evolved activation function, but no other changes to the architecture are made.  The network is optimized using stochastic gradient descent with Nesterov momentum 0.9.  The network is trained for 200 epochs; the initial learning rate is 0.1, and it is decreased by a factor of 0.2 after epochs 60, 120, and 160.  Dropout probability is set to 0.3, and L2 regularization of 0.0005 is applied to the weights.  Data augmentation includes featurewise center, featurewise standard deviation normalization, horizontal flip, and random $32 \times 32$ crops of images padded with four pixels on all sides.  This setup was chosen to mirror the original WRN setup \citep{zagoruyko2016wide} as closely as possible.

During evolution of activation functions, the training is compressed to save time. The network is trained for only 100 epochs; the learning rate begins at 0.1 and is decreased by a factor of 0.2 after epochs 30, 60, and 80.  Empirically, the accuracy achieved by this shorter schedule is sufficient to guide evolution; the computational cost saved by halving the time required to evaluate an activation function can then be used to search for additional activation functions.

\subsection{Residual Network (ResNet-v1-56)}

As with WRN-10-4, when measuring final performance with ResNet-v1-56, the only change to the architecture is replacing the ReLU activations with an evolved activation function.  The network is optimized with stochastic gradient descent and momentum 0.9.  Dropout is not used, and L2 regularization of 0.0001 is applied to the weights.  In the original ResNet experiments \citep{he2016deep}, an initial learning rate of 0.01 was used for 400 iterations before increasing it to 0.1, and further decreasing it by a factor of 0.1 after 32K and 48K iterations.  An iteration represents a single forward and backward pass over one training batch, while an epoch consists of training over the entire training dataset.  In this paper, the learning rate schedule is implemented by beginning with a learning rate of 0.01 for one epoch, increasing it to 0.1, and then decreasing it by a factor of 0.1 after epochs 91 and 137.  (For example, (48K iterations / 45K training images) * batch size of 128 $\approx$ 137.)  The network is trained for 200 epochs in total.  Data augmentation includes a random horizontal flip and random $32 \times 32$ crops of images padded with four pixels on all sides, as in the original setup \citep{he2016deep}.

When evolving activation functions for ResNet-v1-56, the learning rate schedule is again compressed.  The network is trained for 100 epochs; the initial warmup learning rate of 0.01 still lasts one epoch, the learning rate increases to 0.1, and then decreases by a factor of 0.1 after epochs 46 and 68.  When evolving activation functions, their relative performance is more important than the absolute accuracies they achieve.  The shorter training schedule is therefore a cost-efficient way of discovering high-performing activation functions.

\subsection{Preactivation Residual Network (ResNet-v2-56)}

The full training setup, data augmentation, and compressed learning rate schedule used during evolution for ResNet-v2-56 are all identical to those for ResNet-v1-56 with one exception: with ResNet-v2-56, it is not necessary to warm up training with an initial learning rate of 0.01 \citep{he2016identity}, so this step is skipped.

\subsection{All-CNN-C} When measuring final performance with All-CNN-C, the ReLU activation function is replaced with an evolved one, but the setup otherwise mirrors that of \citet{springenberg2015striving} as closely as possible.  The network is optimized with stochastic gradient descent and momentum 0.9.  Dropout probability is 0.5, and L2 regularization of 0.001 is applied to the weights.  The data augmentation involves featurewise centering and normalizing, random horizontal flips, and random $32 \times 32$ crops of images padded with five pixels on all sides.  The initial learning rate is set to 0.01, and it is decreased by a factor of 0.1 after epochs 200, 250, and 300.  The network is trained for 350 epochs in total.

During evolution of activation functions, the same training setup was used.  It is not necessary to compress the learning rate schedule as was done with the residual networks because All-CNN-C trains more quickly.

\subsection{CIFAR-10} As with CIFAR-100, a balanced validation set was created for CIFAR-10 by randomly selecting 500 images from each class, resulting in a training/validation/test split of 45K/5K/10K images.

\section{Implementation and Compute Requirements}
\label{ap:implementation}

High-performance computing in two clusters is utilized for the experiments. One cluster uses HTCondor \citep{thain2005distributed} for scheduling jobs, while the other uses the Slurm workload manager.  Training is executed on GeForce GTX 1080 and 1080 Ti GPUs on both clusters.  When a job begins executing, a parent activation function is selected by sampling $S=16$ functions from the $P=64$ most recently evaluated activation functions.  This is a minor difference from the original regularized evolution \citep{real2019regularized}, which is based on a strict sliding window of size $P$.  This approach may give extra influence to some activation functions, depending on how quickly or slowly jobs are executed in each of the clusters.  In practice the method is highly effective; it allows evolution to progress quickly by taking advantage of extra compute when demand on the clusters is low. 

It is difficult to know ahead of time how computationally expensive the evolutionary search will be.  Some activation functions immediately result in an undefined loss, causing training to end.  In that case only a few seconds have been spent and another activation function can immediately be evaluated.  Other activation functions train successfully, but their complicated expressions result in longer-than-usual training times.  
Although substantial, the computational cost is negligible compared to the cost in human labor in designing activation functions.  Evolution of parametric activation functions requires minimal manual setup and delivers automatic improvements in accuracy.

\section{Baseline Activation Function Details}
\label{ap:baseline}

The following activation functions were used as baseline comparisons in Table \ref{tab:results}.  Some functions were also utilized in the search space (Table \ref{tab:searchspace}).
\begin{itemize}
    \item $\textrm{ReLU} = \max\{x, 0\}$ \citep{nair2010rectified}.
    
    \item $\textrm{ELiSH} = \frac{x}{1+e^{-x}} \texttt{ if } x \geq 0 \texttt{ else } \frac{e^x - 1}{1 + e^{-x}}$  \citep{basirat2018quest}. 
    
    \item $\textrm{ELU} = x \texttt{ if } x \geq 0 \texttt{ else } \alpha (e^x-1)$, with $\alpha = 1$ \citep{elu}.
    
    \item $\textrm{GELU} = x \Phi(x)$, with $\Phi(x) = P(X \leq x), X \sim \mathcal{N}(0, 1)$, approximated as $0.5x(1 + \tanh[\sqrt{2/\pi}(x + 0.044715x^3)])$ \citep{hendrycks2016gaussian}.
        
    \item $\textrm{HardSigmoid} = \max\{0, \min\{1, 0.2x + 0.5\}\}$.
    
    \item $\textrm{Leaky ReLU} = x \texttt{ if } x \geq 0 \texttt{ else } 0.01 x$ \citep{maas2013rectifier}.
        
    \item $\textrm{Mish} = x \cdot \tanh(\textrm{Softplus}(x))$ \citep{misra2019mish}.         
    
    \item $\textrm{SELU} = \lambda x \texttt{ if } x \geq 0 \texttt{ else } \lambda \alpha (e^x-1)$, with $\lambda = 1.05070098$, $\alpha = 1.67326324$ \citep{selu}.
    
    \item $\textrm{sigmoid} = (1 + e^{-x})^{-1}$.
    
    \item $\textrm{Softplus} = \log(e^x + 1)$.
    
    \item $\textrm{Softsign} = x / (|x| + 1)$. 
    
    \item $\textrm{Swish} = x \cdot \sigma(x)$, with $\sigma(x) = (1+e^{-x})^{-1}$ \citep{DBLP:conf/iclr/RamachandranZL18, elfwing2018sigmoid}.
    
    \item $\textrm{tanh} = \frac{e^x - e^{-x}}{e^x + e^{-x}}$.
    
    \item $\textrm{PReLU} = x \texttt{ if } x \geq 0 \texttt{ else } \alpha x$, where $\alpha$ is a per-neuron learnable parameter initialized to 0.25 \citep{he2015delving}.
    
    \item $\textrm{PSwish} = x \cdot \sigma (\beta x)$, where $\beta$ is a per-channel learnable parameter \citep{DBLP:conf/iclr/RamachandranZL18}.
    
    \item $\textrm{APL} = \max\{0, x\} + \sum_{s=1}^S a_s \max\{0, -x+b_s\}$, where $S=7$ and $a_s$ and $b_s$ are per-neuron learnable parameters \citep{apl-agostinelli2014learning}.
    
    \item $\textrm{PAU} = \frac{\sum_{j=0}^m a_j x^j}{1 + |\sum_{k=1}^n b_kx^k|}$, where $m=5, n=4$, and $a_j$ and $b_k$ are per-layer learnable parameters initialized so that the function approximates Leaky ReLU with a slope of 0.01 \citep{pade-molina2019pad}.
        
    \item $\textrm{SPLASH} = \sum_{s=1}^{(S+1)/2} a_s^+ \max\{0, x-b_s\} + a_s^-\max\{0, -x-b_s\}$, where $S=7, b = [0, 1, 2, 2.5]$, and $a_s^+$ and $a_s^-$ are per-layer learnable parameters initialized as $a_1^+=1$ and all other $a = 0$ \citep{tavakoli2020splash}.
    
\end{itemize}

\section{Training with Custom Activation Functions}
\label{ap:custom}

This section demonstrates how to implement different activation functions in a TensorFlow neural network.  For example, the code to create the All-CNN-C architecture with a custom activation function is:

{\scriptsize
\begin{verbatim}
def all_cnn_c(args):
    inputs = Input((32, 32, 3))
    x = Conv2D(96, kernel_size=3, strides=(1, 1), padding='same', kernel_regularizer=l2(0.001))(inputs)
    x = CustomActivation(args)(x)
    x = Conv2D(96, kernel_size=3, strides=(1, 1), padding='same', kernel_regularizer=l2(0.001))(x)
    x = CustomActivation(args)(x)
    x = Conv2D(96, kernel_size=3, strides=(2, 2), padding='same', kernel_regularizer=l2(0.001))(x)
    x = Dropout(0.5)(x)

    x = Conv2D(192, kernel_size=3, strides=(1, 1), padding='same', kernel_regularizer=l2(0.001))(x)
    x = CustomActivation(args)(x)
    x = Conv2D(192, kernel_size=3, strides=(1, 1), padding='same', kernel_regularizer=l2(0.001))(x)
    x = CustomActivation(args)(x)
    x = Conv2D(192, kernel_size=3, strides=(2, 2), padding='same', kernel_regularizer=l2(0.001))(x)
    x = CustomActivation(args)(x)
    x = Dropout(0.5)(x)

    x = Conv2D(192, kernel_size=3, strides=(1, 1), padding='same', kernel_regularizer=l2(0.001))(x)
    x = CustomActivation(args)(x)
    x = Conv2D(192, kernel_size=1, strides=(1, 1), padding='valid', kernel_regularizer=l2(0.001))(x)
    x = CustomActivation(args)(x)
    x = Conv2D(10, kernel_size=1, strides=(1, 1), padding='valid', kernel_regularizer=l2(0.001))(x)
    x = CustomActivation(args)(x)

    x = GlobalAveragePooling2D()(x)
    x = Flatten()(x)
    outputs = Activation('softmax')(x)

    return Model(inputs=inputs, outputs=outputs)
\end{verbatim}}

Here, \texttt{CustomActivation()} is a wrapper that resolves to different activation functions depending on the \texttt{args} parameter.  After importing \texttt{from tensorflow.keras.layers import Activation}, built-in activation functions can be implemented as \texttt{Activation('relu')} or \texttt{Activation('tanh')}, for example.  Activation functions that are not built in and do not contain learnable parameters can be implemented with lambda functions.  For example, the Mish activation function can be implemented with \texttt{Activation(lambda x : x * tf.math.tanh(tf.keras.activations.softplus(x)))}.  Finally, activation functions with learnable parameters simply require subclassing a \texttt{Layer} object.  The code used to implement the PAU activation function is below; APL and SPLASH are implemented in a similar manner.

{\scriptsize
\begin{verbatim}
"""
Padé Activation Units: End-to-end Learning of Flexible Activation Functions in Deep Networks
https://arxiv.org/abs/1907.06732

PAU of degree (5, 4) initialized to approximate Leaky ReLU (0.01)
"""
import tensorflow as tf

from tensorflow.keras.layers import Layer
from tensorflow.keras.initializers import Constant

class PAU(Layer):
    def __init__(self, num_init=None, denom_init=None, param_shape='per-layer', **kwargs):
        super(PAU, self).__init__(**kwargs)
        self.num_init = num_init if num_init else [0.02979246, 0.61837738, 2.32335207, 3.05202660, 1.48548002, 0.25103717]
        self.denom_init = denom_init if denom_init else [1.14201226, 4.39322834, 0.87154450, 0.34720652]
        self.num_weights = []
        self.denom_weights = []
        self.param_shape = param_shape

    def build(self, input_shape):
        if self.param_shape == 'per-layer':
            param_shape = (1,)
        elif self.param_shape == 'per-channel':
            param_shape = list(input_shape[-1:])
        else:
            assert self.param_shape == 'per-neuron'
            param_shape = list(input_shape[1:])

        for i in range(6):
            self.num_weights.append(
                self.add_weight(
                    name=f'a{i}',
                    shape=param_shape,
                    initializer=Constant(self.num_init[i]),
                    trainable=True))
        for i in range(4):
            self.denom_weights.append(
                self.add_weight(
                    name=f'b{i+1}',
                    shape=param_shape,
                    initializer=Constant(self.denom_init[i]),
                    trainable=True))

    def call(self, inputs):
        num = tf.add_n([self.num_weights[i] * tf.math.pow(inputs, i) for i in range(6)])
        denom = 1 + tf.math.abs(tf.add_n([self.denom_weights[i] * tf.math.pow(inputs, i+1) for i in range(4)]))
        return num / denom

    def get_config(self):
        config = super(PAU, self).get_config()
        config.update({'num_init'    : self.num_init,
                       'denom_init'  : self.denom_init,
                       'param_shape' : self.param_shape})
        return config

\end{verbatim}
}

\section{Scope of PANGAEA Search Space}
\label{ap:proofs}
This section shows that any piecewise real analytic function can be represented as a PANGAEA computation graph containing operators from Table \ref{tab:searchspace}.  In the main text, PANGAEA computation graphs were restricted to having at most seven nodes and three learnable parameters $\alpha$, $\beta$, and $\gamma$ for efficiency.  Throughout this section the node and parameter constraints are removed.  Parameters take on the role of any real-valued constant, and the set of functions in PANGAEA without node or parameter constraints is denoted as $\mathcal{G}_\infty$.  Before proving the main result, the following two lemmas are needed.

\begin{lemma}
\label{lemma:real_analytic}
If $f \in C^\omega$ is a real analytic function, then $f \in \mathcal{G}_\infty$.
\end{lemma}
\begin{proof}
As $f$ is real analytic, it can be expressed in the form
\begin{equation}
    \label{eq:real_analytic}
    f(x) = \sum_{n=0}^\infty a_n(x-x_0)^n,
\end{equation}
with parameters $x_0, a_0, a_1, \ldots \in \mathbb{R}$.  As PANGAEA contains the zero, one, addition, and negation operators, the set of integers $\mathbb{Z}$ is contained in $\mathcal{G}_\infty$.  This accounts for the exponent $n$ in the expression above.  All other operators (addition, subtraction, multiplication, exponentiation) in Equation \ref{eq:real_analytic} are included in Table \ref{tab:searchspace}, and so $f \in \mathcal{G}_\infty$.
\end{proof}

\begin{lemma}
\label{lemma:indicator}
Given parameters $a, b \in \mathbb{R}$ where $a < b$, the indicator functions
\begin{align}
    \mathbf{1}_{(-\infty, b)}(x) &= 
    \begin{cases}
        1 & x < b \\
        0 & x \geq b
    \end{cases}\\
    \mathbf{1}_{(a, \infty)}(x) &= 
    \begin{cases}
        1 & x > a \\
        0 & x \leq a
    \end{cases}\\
    \mathbf{1}_{(a,b)}(x) &= 
    \begin{cases}
        1 & x \in (a, b)\\
        0 & x \notin (a, b)
    \end{cases}\\
    \mathbf{1}_a(x) &= 
    \begin{cases}
        1 & x = a \\
        0 & x \neq a
    \end{cases}
\end{align}
are in $\mathcal{G}_\infty$.
\end{lemma}
\begin{proof}
Recall that PANGAEA implements the binary division operator $x_1 / x_2$ as \texttt{tf.math.divide\_no\_nan}, which returns $0$ if $x_2 = 0$.  The indicator function $\mathbf{1}_{(-\infty, b)}(x)$ can then be implemented as
\begin{equation}
    \label{eq:first_indicator}
    \mathbf{1}_{(-\infty, b)}(x) = \frac{\max\{b-x,0\}}{b-x}.
\end{equation}
There are three cases: if $x < b$, the expression evaluates to one.  If $x = b$ or $x > b$, the expression evaluates to zero.  By the same reasoning,
\begin{equation}
    \mathbf{1}_{(a, \infty)}(x) = \frac{\min\{a-x,0\}}{a-x},
\end{equation}
which evaluates to one if $x > a$ and zero otherwise.  Finally, note that
\begin{equation}
    \mathbf{1}_{(a,b)} = \mathbf{1}_{(-\infty, b)}\mathbf{1}_{(a, \infty)}
\end{equation}
and \begin{equation}
    \label{eq:last_indicator}
    \mathbf{1}_a = (1-\mathbf{1}_{(-\infty, a)})(1-\mathbf{1}_{(a, \infty)}).
\end{equation}
All operators in Equations \ref{eq:first_indicator}-\ref{eq:last_indicator} (maximum, minimum, subtraction, multiplication, division, zero, one) are in the PANGAEA search space in Table \ref{tab:searchspace}, and so the indicator functions are in $\mathcal{G}_\infty$.
\end{proof}

\begin{theorem}
\label{thm:piecewise_real_analytic}
If a function $f$ is piecewise real analytic, then $f \in \mathcal{G}_\infty$.
\end{theorem}
\begin{proof}
If a function $f$ is piecewise real analytic, then it is representable by the form
\begin{equation}
    f(x) = 
    \begin{cases}
    f_0(x) & x \in (-\infty, k_1) \\
    K_1 & x = k_1\\
    f_1(x) & x \in (k_1, k_2) \\
    K_2 & x = k_2 \\
    & \vdots \\
    f_{n-1}(x) & x \in (k_{n-1}, k_n) \\
    K_n & x = k_n \\
    f_n(x) & x \in (k_n, \infty)
    \end{cases},
\end{equation}
where parameters $K_1, K_2, \ldots, K_N, k_1, k_2, \ldots, k_n \in \mathbb{R}$ are real-valued, $k_1 < k_2 < \cdots < k_n$ are increasing, and $f_0, f_1, \ldots, f_n \in C^\omega$ are real analytic functions.  An equivalent representation of $f$ is the following:
\begin{equation}
    \label{eq:alt_representation}
    \begin{aligned}
        f(x) = \mathbf{1}_{(-\infty, k_1)}(x)f_0(x) + \mathbf{1}_{k_1}(x)K_1 + \mathbf{1}_{(k_1,k_2)}(x)f_1(x) + \mathbf{1}_{k_2}(x)K_2 + \cdots \\ \phantom{} + \mathbf{1}_{(k_{n-1},k_n)}(x)f_{n-1}(x) + \mathbf{1}_{k_n}(x)K_n + \mathbf{1}_{(k_n,\infty)}(x)f_{n}(x).
    \end{aligned}
\end{equation}
By Lemmas \ref{lemma:real_analytic} and \ref{lemma:indicator}, the real analytic functions $f_i$ and the indicator functions $\mathbf{1}_{(\cdot, \cdot)}$ are in $\mathcal{G}_\infty$.  Beyond these functions, Equation \ref{eq:alt_representation} utilizes only addition and multiplication, both of which are operators included in Table \ref{tab:searchspace}.  Therefore, $f \in \mathcal{G}_\infty$.
\end{proof}

\section{Size of the PANGAEA Search Space}
\label{ap:searchspace}

This section analyzes the size of the PANGAEA search space as implemented in experiments in the main text.  Let $g$ represent a general computation graph, and let $f$ represent a specific activation function representable by $g$.  For example, if we have $g(x) = \texttt{binary}(\texttt{unary1}(x), \texttt{unary2}(x))$, then one possible activation function is $f(x) = \tanh(x) + \textrm{erf}(x)$, and another could be $f(x) = \alpha| x| \cdot \sigma(\beta \cdot x)$.  

Let $U=27$ and $B=7$ be the number of unary and binary operators in the PANGAEA search space, respectively, and let $E=3$ be the maximum number of learnable parameters that can be used to augment a given activation function.  Given a computation graph $g$, let $u_g$ and $b_g$ be the number of unary and binary nodes and let $e_g$ be the number of edges in $g$.  For example, with the functional form $g(x) = \texttt{unary1}(\texttt{unary2}(x))$, we have $u_g=2$, $b_g=0$, and $e_g=3$.  With $g(x) = \texttt{binary}(\texttt{unary1}(x), \texttt{unary2}(x))$, we have $u_g=2$, $b_g=1$, and $e_g=5$.  The quantity $e_g$ includes the edges from the input nodes $x$ and edges to the output node $g(x)$ (see Figure 1 in the main text).  

\begin{table}
    \centering
    \begin{tabular}{cccccc}
    \toprule
    & Binary Nodes $b_g$ & Unary Nodes $u_g$ & Edges $e_g$ & Arrangements & Activation Functions \\ \midrule
    $\mathcal{G}_1$                & 0 & 1 & 2 & 1  & 108 \\ \midrule
    $\mathcal{G}_2$                & 0 & 2 & 3 & 1 & 5,832 \\ \midrule
    \multirow{2}*{$\mathcal{G}_3$} & 0 & 3 & 4 & 1 & \multirow{2}*{427,923} \\
                                   & 1 & 2 & 5 & 1 \\ \midrule
    \multirow{2}*{$\mathcal{G}_4$} & 0 & 4 & 5 & 1 & \multirow{2}*{31,177,872} \\
                                   & 1 & 3 & 6 & 3 \\ \midrule
    \multirow{3}*{$\mathcal{G}_5$} & 0 & 5 & 6 & 1 & \multirow{3}*{2,210,558,364} \\
                                   & 1 & 4 & 7 & 6 \\
                                   & 2 & 3 & 8 & 2 \\ \midrule
    \multirow{3}*{$\mathcal{G}_6$} & 0 & 6 & 7 & 1 & \multirow{3}*{152,059,087,566} \\
                                   & 1 & 5 & 8 & 10 \\
                                   & 2 & 4 & 9 & 10 \\ \midrule
    \multirow{4}*{$\mathcal{G}_7$} & 0 & 7 & 8 & 1 & \multirow{4}*{10,015,741,690,785} \\
                                   & 1 & 6 & 9 & 15 \\
                                   & 2 & 5 & 10 & 30 \\
                                   & 3 & 4 & 11 & 1 \\
    \bottomrule
    \end{tabular}
    \caption{The number of activation functions representable by a computation graph with a given number of nodes.  The PANGAEA search space contains over ten trillion activation functions, and therefore provides a good foundation for finding powerful activation functions with different properties.}
    \label{tab:pangaea_analysis}
\end{table}

Let $\mathcal{F}_g$ denote the set of all activation functions $f$ that can be represented within the computation graph $g$.  By enumerating the different choices of unary and binary operators, as well as the locations for up to three learnable parameters $\alpha$, $\beta$, and $\gamma$, we find the size of the set to be
\begin{equation}
    | \mathcal{F}_g | = U^{u_g} \cdot B^{b_g} \cdot \sum_{i=0}^E \binom{e_g}{i}.
\end{equation}
Let $\mathcal{G}_j$ denote the set of computation graphs $g$ containing $j$ nodes.  For example,
\begin{equation}
\mathcal{G}_3 = \{ g(x) = \texttt{unary1}(\texttt{unary2}(\texttt{unary3}(x))), g(x) = \texttt{binary}(\texttt{unary1}(x), \texttt{unary2}(x)) \}.    
\end{equation}
Table \ref{tab:pangaea_analysis} shows the possible combinations of binary nodes $b_g$, unary nodes $u_g$, and edges $e_g$ for each set $\mathcal{G}_j$.  Additionally, the table shows the number of computation graph arrangements possible for a given $b_g$, $u_g$, and $e_g$.  For example, if $b_g = 2$, $u_g = 3$, and $e_g = 8$, the computation graph could take one of two forms: either 
\begin{equation}
g(x) = \texttt{binary1}(\texttt{binary2}(\texttt{unary1}(x), \texttt{unary2}(x)), \texttt{unary3}(x))    
\end{equation}
or
\begin{equation}
g(x) = \texttt{binary1}(\texttt{unary1}(x), \texttt{binary2}(\texttt{unary2}(x), \texttt{unary3}(x))).
\end{equation}
The number of activation functions in PANGAEA is therefore
\begin{equation}
    \sum_{j=1}^7 \sum_{g \in \mathcal{G}_j} |\mathcal{F}_g| = 10{,}170{,}042{,}948{,}450.
\end{equation}

Naturally there exist duplicates within this space.  The functions $f(x) = \textrm{ReLU}(x)$ and $f(x) = \max\{x, 0\}$ have different computation graphs but are functionally identical.  Nevertheless, this analysis still provides a useful characterization of the size and diversity of the PANGAEA search space.  It is orders of magnitude larger than spaces considered in prior work \citep{bingham2020gecco, DBLP:conf/iclr/RamachandranZL18, basirat2018quest}, and yet PANGAEA consistently discovers functions that outperform ReLU and other baseline functions.

\section{Additional Results with Learnable Activation Functions}
\label{ap:pausplash}

PAU and SPLASH achieved worse-than-expected performance in Table \ref{tab:results}, so additional experiments were run to investigate their behavior.

\paragraph{PAU}

\citet{pade-molina2019pad} utilized a specialized training setup to achieve their results with PAU.  In particular, they used a constant learning rate and no weight decay for the PAU layers, but used a learning rate decay of 0.985 per epoch and weight decay of $0.0005$ for the other weights.  They also used a smaller batch size of 64, and trained for 400 epochs instead of 200.  Even though the paper does not mention it, it is possible that such a specialized setup is necessary to achieve good performance with PAU.  The experiments in this appendix utilized this same training setup (but only trained for 200 epochs for fairness) to verify that the PAU implementation was correct.  

Unfortunately, some relevant hyperparameters were not included in the PAU paper \cite{pade-molina2019pad}.  These settings include the fixed learning rate used for the PAU layers, whether Nesterov momentum is utilized, and which approximation of Leaky ReLU is used to initialize the PAU weights.  This missing information makes it difficult to replicate the original performance exactly.  After significant trial-and-error, the following settings worked well: The learning rate for the PAU layers was 0.01, the initial learning rate for other weights was also 0.01, Nesterov momentum was not used, and the PAU weights were initialized to approximate Leaky ReLU with a slope of 0.01. 

Table \ref{tab:pau_specialized_setup} shows the performance of WRN-10-4 and ResNet-v2-56 using these discovered hyperparameters and the specialized training setup from \citet{pade-molina2019pad}.  The performance is comparable to other baseline activation functions.  In some cases, the runs failed because of training instability (results were filtered out if the training accuracy was below 0.5).  For all hyperparameter combinations tested, PAU was unstable with ResNet-v1-56.  Thus, it is possible to get good performance with PAU, but the performance is highly sensitive to the training setup and choice of hyperparameters. 

\begin{table}
    \centering
    \begin{tabular}{lll}
    \toprule
    & \textbf{WRN-10-4} & \textbf{ResNet-v2-56} \\
    \midrule
    Accuracy & $62.56 {\scriptscriptstyle \pm 4.84}$ & 
    $69.59 {\scriptscriptstyle \pm 1.66}$\\
    Failed Runs & 3 of 10 & 7 of 10 \\
    \bottomrule
    \end{tabular}
    \caption{CIFAR-100 test accuracy with PAU using a specialized training setup.  Performance is comparable to other baseline activation functions, but some runs fail due to training instability.}
    \label{tab:pau_specialized_setup}
\end{table}

Note that a standard, most commonly used setup was used throughout the main experiments in the paper for all baseline comparisons. The reason is that there are dozens of such comparisons in this paper, and it is possible that each one could benefit from a specialized setup---a setup that may not even be fully known at this time. Therefore, a standard setup was necessary to ensure that the comparisons are fair.

\paragraph{SPLASH}

\begin{figure}[t]
    \centering
    \includegraphics[width=0.49\textwidth]{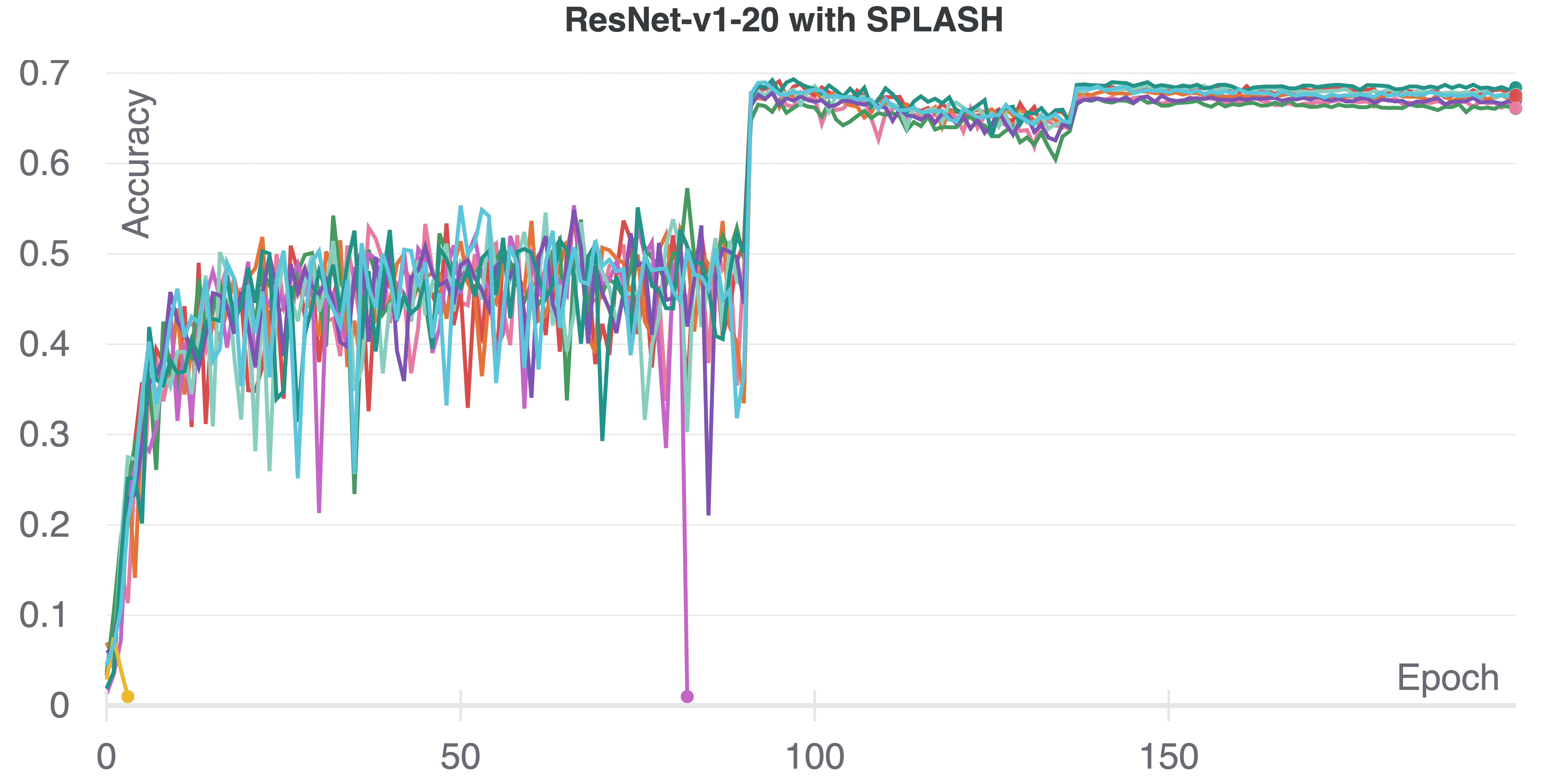}
    \includegraphics[width=0.49\textwidth]{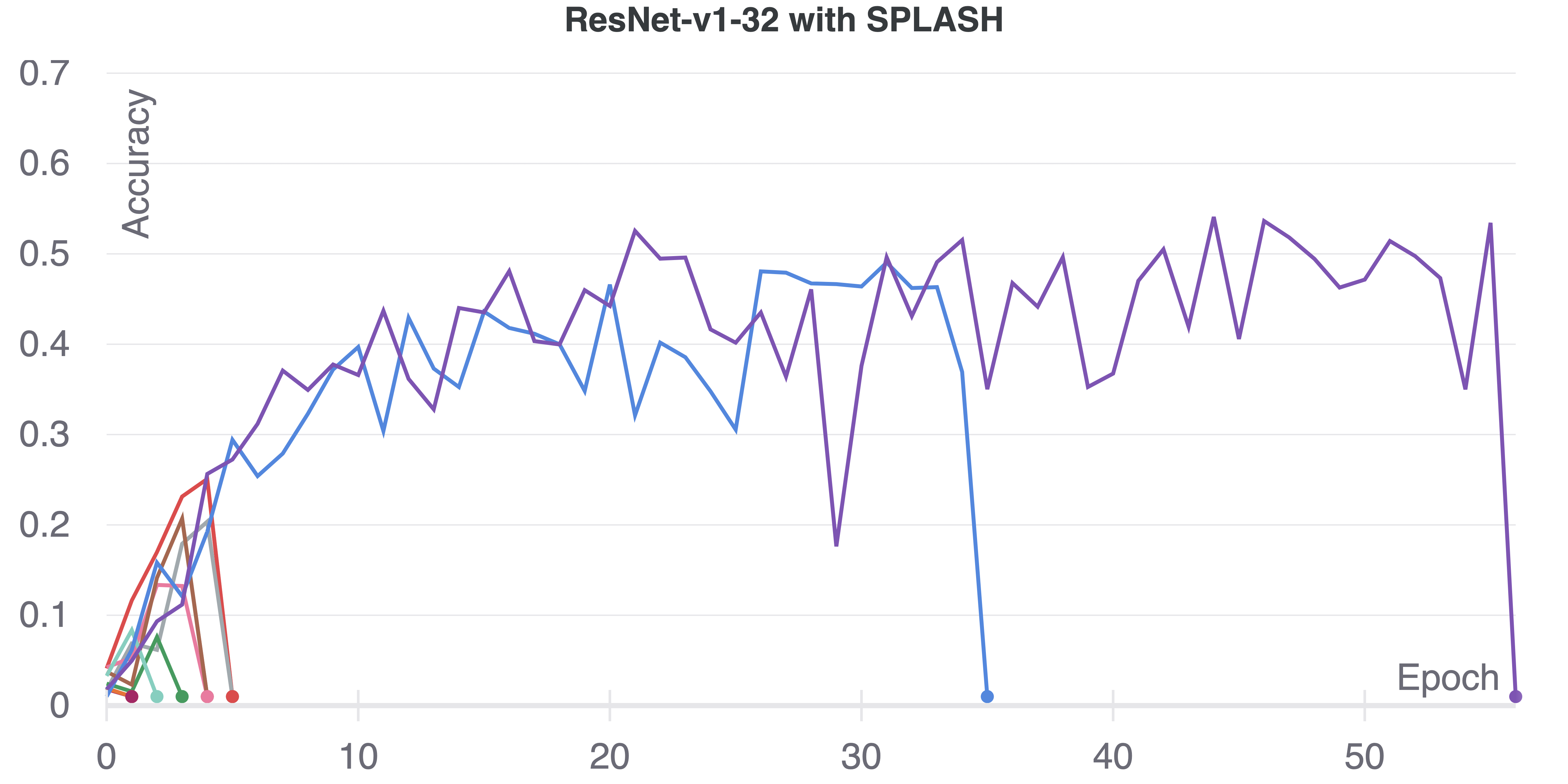}
    \caption{ResNet-v1-20 and ResNet-v1-32 test accuracy on CIFAR-100 with the SPLASH activation function.  SPLASH units work well in shallow networks, but become unstable with increased depth.  This result explains the success of SPLASH in the original work by \citet{tavakoli2020splash}, and also shows why SPLASH units fail with the architectures considered in this work.}
    \label{fig:rnv1_splash}
\end{figure}

In addition to ResNet-v1-56 in the main text, SPLASH was trained with ResNet-v1-20, ResNet-v1-32, and ResNet-v1-44 for a more thorough characterization of its performance.  Each architecture was trained ten times, resulting in training curves shown in Figure \ref{fig:rnv1_splash}.

With ResNet-v1-20, final test accuracy of the independent runs is between 0.661 and 0.684, which agrees with the results by \citet{tavakoli2020splash}.  However, two of the ten runs failed in the middle of training because the loss became undefined (Figure \ref{fig:rnv1_splash}), suggesting that SPLASH units can be unstable.  Progressing to the deeper ResNet-v1-32, the effect was more pronounced.  As shown in Figure \ref{fig:rnv1_splash}, only two of the ten runs progressed past epoch 30, while no run trained to completion.  With ResNet-v1-44 and ResNet-v1-56, training failed within the first epoch, so the training curves are not shown.

These results thus confirm that the implementation is correct, reproducing the results of \citet{tavakoli2020splash}. However, they also lead to the interesting observation that SPLASH units are effective with shallow networks but struggle with deeper ones, like the ones evaluated in this paper.

\end{document}